\documentclass[runningheads]{llncs}
\usepackage{graphicx}

\usepackage{tikz}
\usepackage{comment}
\usepackage{amsmath,amssymb} 
\usepackage{amscd}
\usepackage[bottom]{footmisc}
\usepackage{multirow}
\usepackage{color}
\usepackage[pagebackref,breaklinks,colorlinks]{hyperref}
\usepackage{caption}
\usepackage{subcaption}
\usepackage{floatrow}
\newfloatcommand{capbtabbox}{table}[][\FBwidth]
\usepackage[accsupp]{axessibility}  

\def\etal{\textit{et al.}}

\usepackage{booktabs}
\def\etal{\textit{et al.}}
\usepackage[utf8]{inputenc}
\usepackage{pifont}
\newcommand{\cmark}{\ding{51}}%
\newcommand{\xmark}{\ding{55}}
\newcommand{\R}{{\mathbb R}}

\usepackage[capitalize]{cleveref}
\crefname{section}{Sec.}{Secs.}
\Crefname{section}{Section}{Sections}
\Crefname{table}{Table}{Tables}
\crefname{table}{Tab.}{Tabs.}

\begin{document}
\pagestyle{headings}
\mainmatter
\def\ECCVSubNumber{6963}  

\title{3D Shape Sequence of Human Comparison and Classification  using Current and Varifolds} 

\titlerunning{3D Shape Sequence of Human Comparison  using Current and Varifolds}
%
\author{Emery Pierson\inst{1}
\and Mohamed Daoudi\inst{1,2}
\and Sylvain Arguillere \inst{3}}
\authorrunning{E. Pierson et al.}
%
\institute{Univ. Lille, CNRS, Centrale Lille, UMR 9189 CRIStAL, F-59000 Lille, France 
\email{emery.person@univ-lille.fr}\\ \and
IMT Nord Europe, Institut Mines-Télécom, Centre for Digital Systems
\email{mohamed.daoudi@imt-nord-europe.fr}\\
\and Univ. Lille, CNRS, UMR 8524 Laboratoire Paul Painlevé, Lille, F-59000, France 
\email{sylvain.arguillere@univ-lille.fr}}
\maketitle

\begin{abstract}
    In this paper we address the task of the comparison and the classification of 3D shape sequences of human. The non-linear dynamics of the human motion and the changing of the surface parametrization over the time make this task very challenging.
    To tackle this issue, we propose to embed the 3D shape sequences in an infinite dimensional space, the space of varifolds, endowed with an inner product that comes from a given positive definite kernel. More specifically, our approach involves two steps: 1) the surfaces are represented as varifolds, this representation induces metrics equivariant to rigid motions and invariant to parametrization; 2) the sequences of 3D shapes are represented by Gram matrices derived from their infinite dimensional Hankel matrices. 
    The problem of comparison of two 3D sequences of human is formulated as a comparison of two Gram-Hankel matrices.
    Extensive experiments on CVSSP3D and Dyna datasets show that our method is competitive with state-of-the-art in 3D human sequence motion retrieval. Code for the experiments is available at~\url{https://github.com/CRISTAL-3DSAM/HumanComparisonVarifolds}
    
    \keywords{3D Shape Sequence \and Varifold \and 3D Shape Comparison \and Hankel matrix}
\end{abstract}


\section{Introduction}
Understanding 3D human shape and motion  has many important applications, such as ergonomic design of products, rapid modeling of realistic human characters for virtual worlds, and an early detection of abnormality in predictive clinical analysis.
Recently, 3D human data has become highly available as a result of the availability of huge MoCap (Motion Capture) datasets~\cite{cmu2018,Aristidou:2014:GRAPP} along with the evolution of 3D human body representation~\cite{SMPL:2015} leaded to the availability of huge artificial human body datasets~\cite{amass,cvssp3d}. 
In the meantime, evolutions in 4D technology for capturing moving shapes lead to paradigms with new multi-view and 4D scan acquisition systems that enable now full 4D models of human shapes that include geometric, motion and appearance information~\cite{tsiminaki:hal-00977755,bogo_dynamic_2017,pons2015dyna,gkalelis2009i3dpost}.  

\begin{figure}
    \centering
    \begin{subfigure}{0.25\linewidth}
        \includegraphics[width=\linewidth]{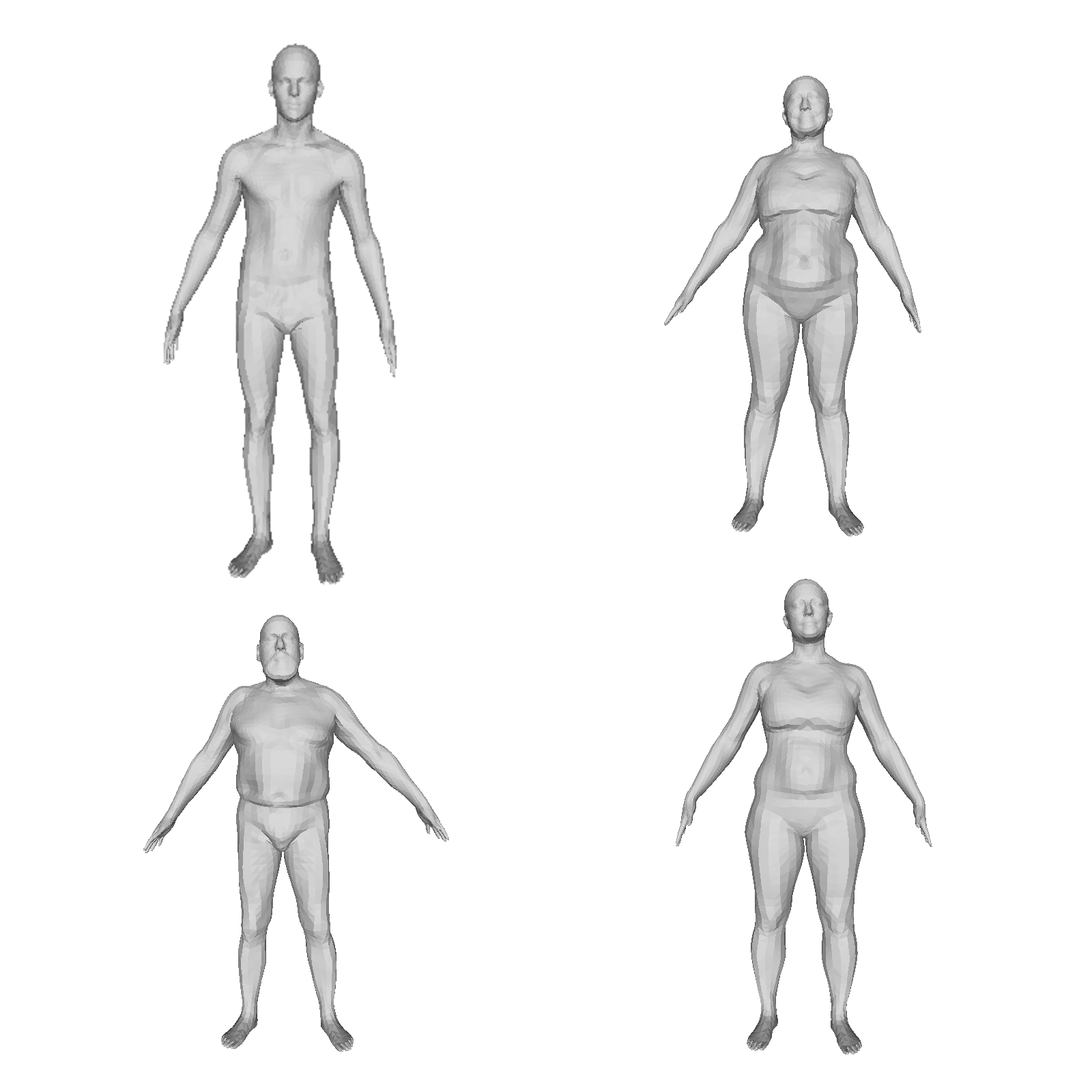}
        \caption{}
    \end{subfigure}
    \begin{subfigure}{0.25\linewidth}
        \includegraphics[width=\linewidth]{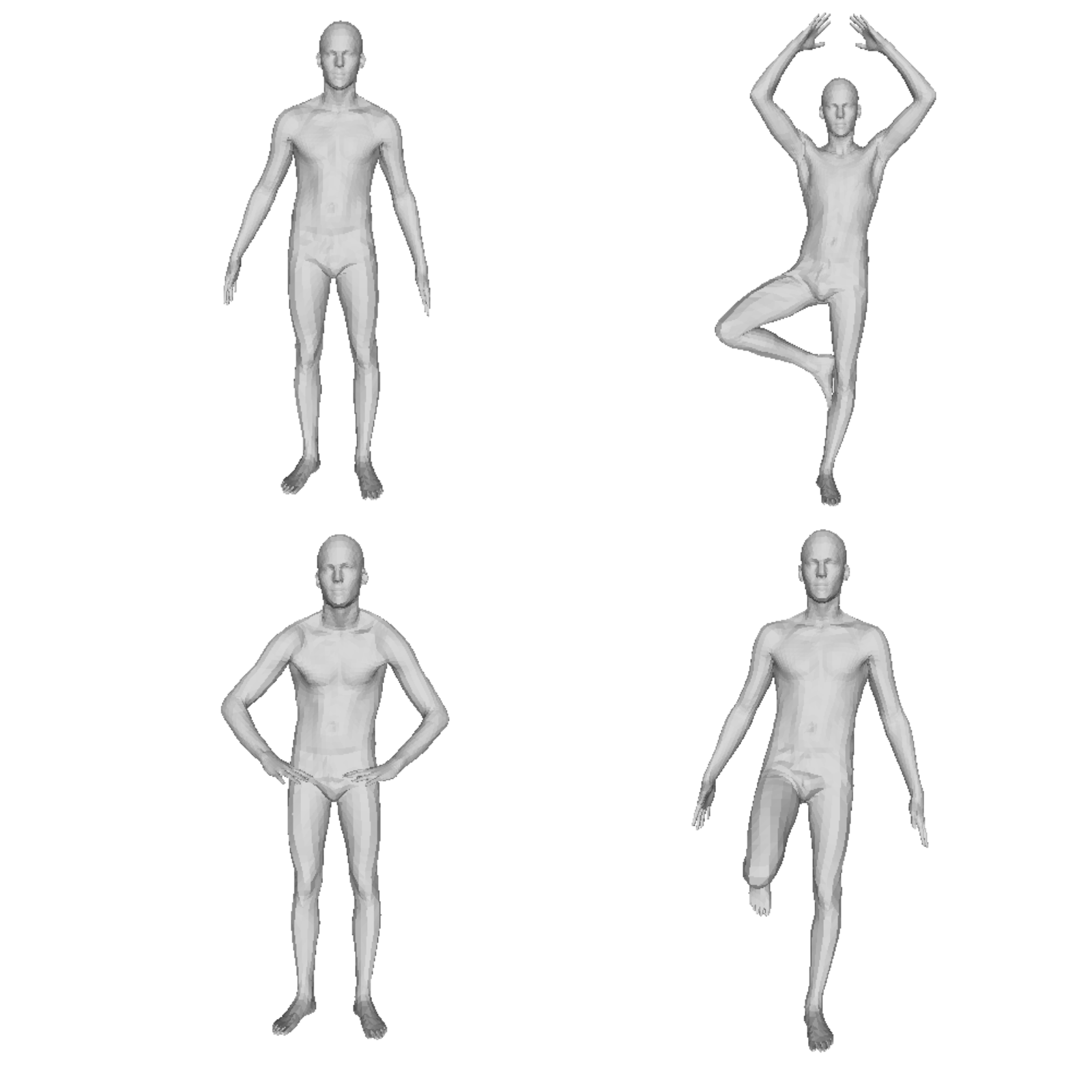}
        \caption{}
    \end{subfigure}
    \begin{subfigure}{0.40\linewidth}
        \includegraphics[width=\linewidth]{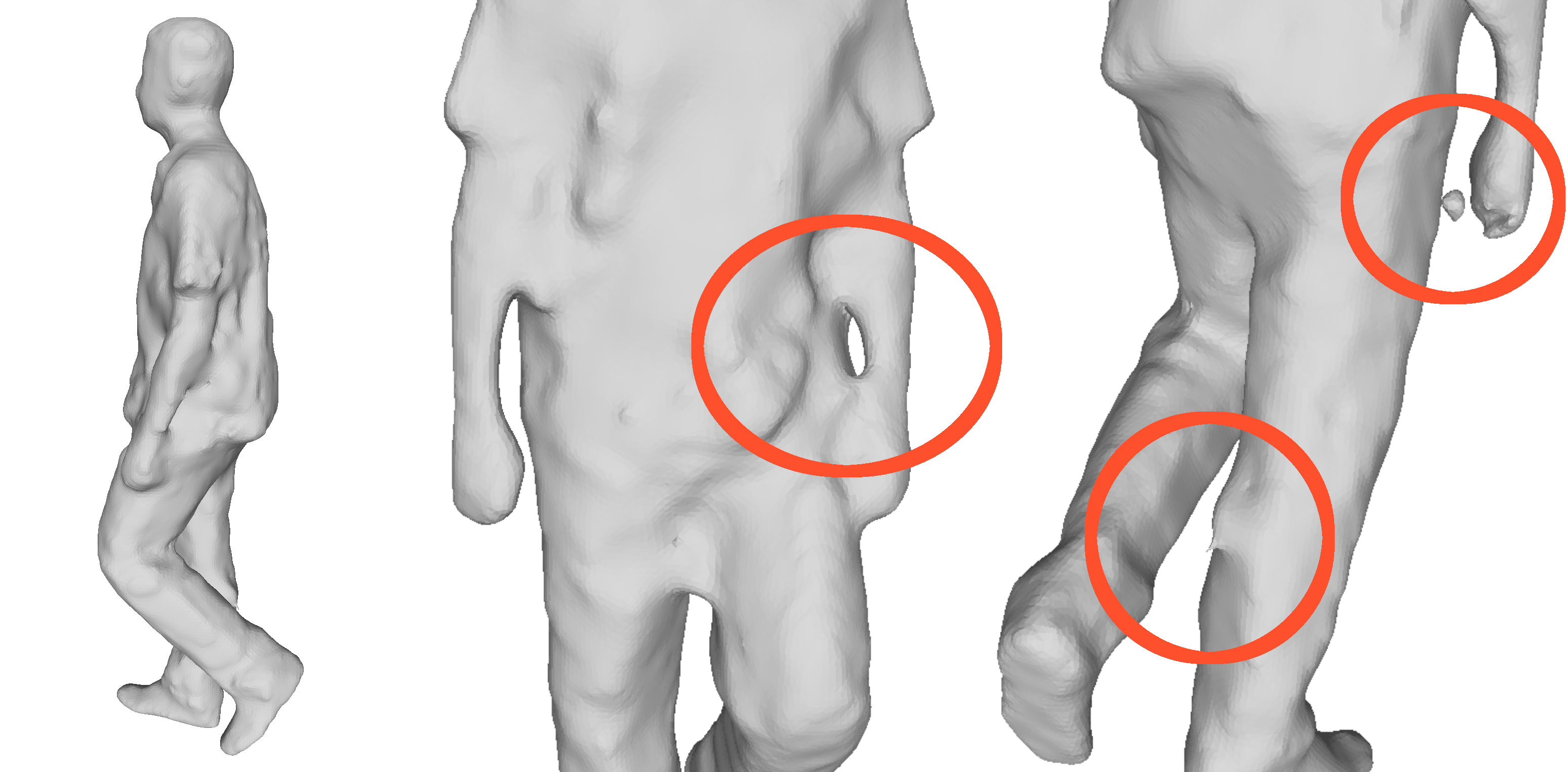}
        \caption{}
    \end{subfigure}
    
    \caption{\textbf{Different challenges of 3D Human Sequence Comparison:} (a) Shape variability within and across shape classes, (b) variability in pose and articulation, (c) noisy and arbitrary mesh parameterizations (topological noise, vertex noise and disconnected components).}
    \label{fig:challenges}
\end{figure}

The first difficulty in analyzing shapes of 3D human comes from noise, variability in pose and articulation, arbitrary mesh parameterizations during data collection, and shape variability within and across shape classes. Some examples of 3D human highlighting these issues are illustrated in~\Cref{fig:challenges}. In particular, the metrics and representations should have certain invariances or robustness to the above-mentioned variability. Recently, Kaltenmark~\etal~\cite{kaltenmark2017general}  have proposed a general framework for 2D and 3D shape similarity measures, invariant to parametrization and equivariant to rigid transformations. More recently, Bauer \etal~\cite{bauer2021numerical}, adopted the varifold fidelity metric as a regularizer for the problem of reparameterization in the framework of elastic shape matching using the SRNF~\cite{JermynEccv2012} representation. Motivated by the progress of using varifolds and current in shape analysis, we propose to compare 3D surface of human shapes by comparing their varifolds.

As a second difficulty, it is critical to identify precise mathematical representations of underlying shapes and then impose efficient dynamical models on representation spaces that capture the essential variability in shape evolutions. In addition to the nonlinearity of shape spaces, one expects nonlinearity in temporal evolutions that makes the inference process difficult.  In this paper, we propose to use Gram matrices derived from Hankels matrices to represent the dynamic of human motion.

\begin{figure}[!t]
    \centering
    \includegraphics[width=0.8\linewidth]{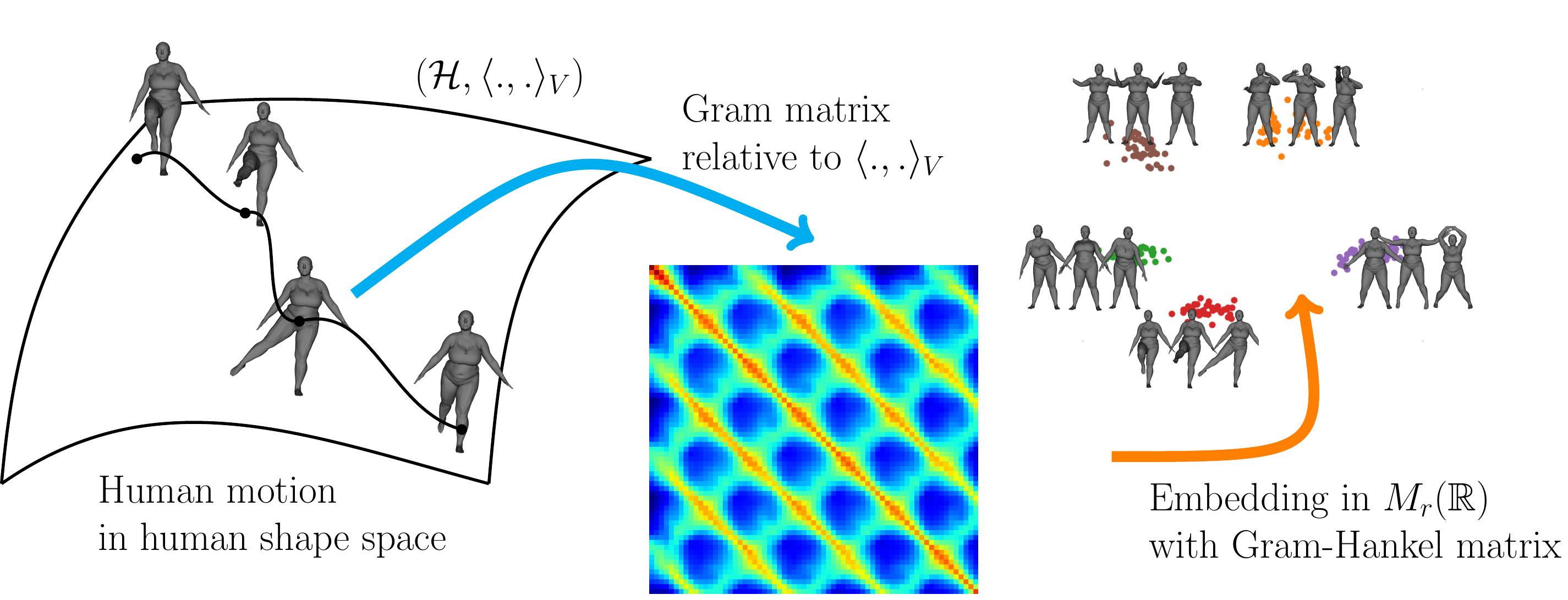}
    \caption{\textbf{Overview of our method}. We embed the human shape space $\mathcal{H}$ in an infinite dimensional Hilbert space with inner product from a positive definite kernel $\langle ., .\rangle_V$ inspired by varifold framework. Using this kernel product we are able to compute Gram matrix relative to a motion. Each of this Gram matrix is transformed to Gram Hankel matrix of size $r$. The Frobenius distance in $M_r(\mathbb{R})$ is used to retrieve similar 3D sequences.}
    \label{Fig:Overview}
\end{figure}

In our approach, as illustrated in Figure~\ref{Fig:Overview}, we propose to embed the human shape space $\mathcal{H}$ in an infinite dimensional Hilbert space with inner product corresponding to a positive definite kernel $\langle ., .\rangle_V$ inspired by the varifold framework. Using this kernel product we are able to compute the Gram matrix relative to a motion. Each of this Gram matrix is transformed to Gram-Hankel matrix of fixed size $r$.

In summary, the main contributions of this article are: \textbf{\emph{(i)}} We represent 3D human surfaces as varifold. This representation is equivariant to rotation and invariant to the parametrization. This representation allows us to define an inner product between two 3D surfaces represented by varifolds.
    \textbf{\emph{(ii)}} It is the first use of the space of varifolds in human shape analysis. The framework does not assume that the correspondences between the surfaces are given. 
     \textbf{\emph{(iii)}} We represent 4D surfaces by Hankel matrices. This key contribution enables the use of standard computational tools
     based on the inner product defined between two varifolds. The dynamic information of a sequence of 3D human shape is encapsulated in Hankel matrices and we propose to compare sequences by using the distance between the resulting Gram-Hankel matrices;   \textbf{\emph{(iv)}} The experiments results show that the proposed approach improves 3D human motion retrieval state-of-the-art  and it is robust to noise.
    
\section{Related Work}

\subsection{3D Human Shape Comparison}
The main difficulty in comparing human shapes of such surfaces is that there is no preferred parameterization that can be used for registering and comparing features across surfaces. Since the shape of a surface is invariant to its parameterization, one would like an approach that yields the same result irrespective of the parameterization. The linear blending approaches~\cite{hasler2009statistical,anguelov2005scape,SMPL:2015} offer a good representation for human shape, along with a model of human deformations while being able to distinguish shape and pose deformations. However these methods need  additional information on the raw scans such as MoCap markers~\cite{hasler2009statistical,anguelov2005scape}, gender of the body, or additional texture information~\cite{SMPL:2015,bogo_dynamic_2017} to retrieve such representations. Recently, deep learning approaches~\cite{tan2018variational,gdvae_2019,zhou20unsupervised} propose human bodies latent spaces that share common properties with linear blending models. However, they require training data with the same mesh parameterization and are sensitive to noise. Moreover, most current techniques treat shape and motion independently, with devoted techniques for either shape or motion in isolation.

Kurtek~\etal~\cite{Kurtekpami2012} and Tumpach~\etal~\cite{TumpachTPAMI2016} propose the quotient of the space of embeddings of a fixed surface $S$ into $\R^3$ by the action of the orientation-preserving diffeomorphisms of $S$ and the group of Euclidean transformations, and provide this quotient with the structure of an infinite-dimensional manifold. The shapes are compared using a Riemannian metric on a \textit{pre-shape space} $\mathcal{F}$ consisting 
of embeddings or immersions of a model manifold into 
the 3D Euclidean space $\mathbb{R}^3$.  Two embeddings correspond to the same shape in $\mathbb{R}^3$ if and only if  they differ by an element of  a shape-preserving transformation group. However the use of these approaches on human shape analysis assume a spherical parameterization of the surfaces.  Pierson~\etal~\cite{Pierson-wacv-2022} propose a Riemannian approach for human shape analysis. This approach provides encouraging results, but it requires the meshes to be registered to a template.
Recently, the framework of varifolds have been presented for application to shape matching. Charon~\etal~\cite{charon2013varifold} generalize the framework of currents, which defines a restricted type of geometric measure on surfaces, by the varifolds framework representing surfaces as a measure on $\mathbb{R}^3 \times \mathbb{S}^2$. The proposed varifolds representation is parameterization invariant, and does not need additional information on raw scans.  Inspired by these recent results, we will demonstrate the first use of this mathematical theory in 3D human shape comparison. 
\subsection{3D Human Sequence Comparison}
A general approach adopted when comparing 3D sequences is the extension of static shape descriptors such as 3D shape distribution, Spin Image, and spherical harmonics to include temporal motion information~\cite{huang2010shape,veinidis2019effective,pierson2021projection}. While these approaches require the extraction of shape descriptors, our approach does not need a 3D shape feature extraction. It is based on the comparison of surface varifolds within a sequence.
In addition, the comparison of 3D sequences require an alignment of the sequences.
The Dynamic Time Warping (DTW) algorithm was defined to match temporally distorted time series, by finding an optimal warping path between time series. It has been used for several computer vision applications~\cite{amor2016action,kacem2018novel} and alignment of 3D human sequences~\cite{veinidis2019effective,pierson2021projection}. However, DTW does not define a proper distance (no triangle inequality). In addition, a temporal filtering is often required for the alignment of noisy meshes~\cite{slama_3d_2014}. Our approach enables the comparison of sequences of different temporal duration, does not need any alignment of sequences and is robust to noisy data. We model a sequence of 3D mesh as a dynamical system. The parameters of the dynamical system are embedded in our Hankel matrix-based representation. Hankel matrices have already been adopted successfully for skeleton action recognition in~\cite{zhang_efficient_2016}. As we do not have finite dimensional features to build such matrix numerically, we define a novel Gram-Hankel matrix, based on the kernel product defined from surface varifold. This matrix is able to model the temporal dynamics of the 3D meshes.

\section{Proposed Method}
\subsection{Comparing 3D Shapes using Geometric Measures}
The varifolds framework is a geometry theory used to solve famous differential geometry problems such as the Plateau's Problem. We invite the interested reader to read an introduction of the theory in~\cite{almgren1966plateau}. We focus here on the work of Charon~\etal~\cite{charon2013varifold}, followed by Kaltenmark~\etal~\cite{kaltenmark2017general} who proposed to use the varifolds framework for discretized curves and surfaces. They designed a fidelity metric using the varifold representation. This fidelity metric is proposed for 2D artificial contour retrieval, and used in 3D diffeomorphic registration in the Large deformation diffeomorphic metric mapping (LDDMM) framework. To our knowledge our work is the first use of such representation for the analysis of human shape. As far as the authors are aware, our work is the first use of such representation for the analysis of human shape. It is also the first use of the space of varifolds purely for itself, as an efficient way to perform direct computations on shapes.

A \textit{varifold} is a measure $\mu$ on $\R^3\times\mathbb{S}^2$. The integral of a function $f:\R^3\times\mathbb{S}^2\rightarrow\R$ with respect to such a measure is denoted $\int_{\R^3\times\mathbb{S}^2}fd\mu$. Given $S$ a smooth compact surface with outer normal unit vector field $x\mapsto n(x)$, the core idea of~\cite{charon2013varifold} is to represent $S$ as a varifold. This is done in practice through the formula $\int_{\R^3\times\mathbb{S}^2}fdS= \int_Sf(x,n(x))dA(x)$, with $dA(x)$ the surface area measure of $S$ at $x$. Now, given a triangulated surface $M$ of a 3D human shape with triangle faces $T_1, ..., T_m$, when the triangles are sufficiently small, each triangular face $T_i$ is represented as an atomic measure $a_i\delta_{c_i,n_i}$ where $c_i$ is the barycenter of the triangulated face, $n_i$ the oriented normal of the face, $a_i$ its area, and $\delta$ representing the dirac mass. The varifold representation of the total shape  $M$ is simply given by the sum of all these measures: $M=\sum_{i=1}^ma_i\delta_{c_i,n_i}$. To illustrate, integrating a function $f$ on $\R^3\times \mathbb{S}^2,$ with respect to $M$ yields $\int_{\R^3\times \mathbb{S}^2} fdM=\sum_{i=1}^m a_if(c_i,n_i)$.

A varifold $\mu$ can be converted into a function $\Phi_\mu$ on $\R^3\times\mathbb{S}^2$ using a \textit{reproducing kernel} that comes from the product of two positive definite kernels: $k_{\textit{pos}}:\mathbb{R}^3\times \mathbb{R}^3\rightarrow\R$, and $k_{\textit{or}}:\mathbb{S}^2\times \mathbb{S}^2\rightarrow\R$. We just define $\Phi_\mu(x,v)=\int_{\R^3\times\mathbb{S}^2}k_{pos}(y,x)k_{or}(w,v)d\mu(y,w).$ For a triangulated surface $M$, we get $\Phi_M(x,v)=\sum_{i=1}^ma_ik_{pos}(c_i,x)k_{or}(n_i,v)$.

One obtains a Hilbert product between any two varifolds $\mu,\nu$ as follows~: $\left<\mu,\nu\right>_V=\left<\Phi_\mu,\Phi_\nu\right>_V=\int \Phi_\mu d\nu=\int\Phi_\nu d\mu$, so that $$\left<\mu,\nu\right>_V=\iint k_{\textit{pos}}(x,y)k_{\textit{or}}(v,w)d\mu(x,v)d\nu(y,w). $$ We deduce the explicit expression for that product between two triangulated 3D shapes $M$ and $N$~:

\begin{equation}
    \langle M, N\rangle_V= \langle \Phi_M, \Phi_N\rangle_V = \sum_{i=1}^{m} \sum_{j=1}^{n} a^M_i a^N_j k_{\textit{pos}}(c^M_i, c^N_j) k_{\textit{or}}(n^M_i, n^N_j)
    \label{eq: kernel_full}
\end{equation}
Where $m, n$ are the number of faces of $M$ and $N$. The continuous version of this product presented in~\cite{charon2013varifold} is parametrization invariant.

An important part of such a product is that it can be made equivariant to rigid transformation by carefully choosing the kernels. First we define how to apply such a deformation on a varifold. Given a rotation $R\in SO(3)$ of $\R^3$ and a vector $T\in \R^3$, the rigid transformation $\phi:x\mapsto Rx+T$ yields the push-forward transformation $\mu\mapsto\phi_\#\mu$ through $\int fd\phi_\#\mu=\int f(Rx+T,Rv)d\mu$ on the space of varifolds. For a triangulated surface $M$, $\phi_\#M$  is just $\phi(M)$, the surface obtained by applying the rigid motion $\phi$ to the surface $M$ itself. We have the following important result~:
\begin{theorem}\label{thm:equi_tri}
If we define the positive definite kernels as following:
\begin{equation*}
    \begin{array}{clc}
        k_{\textit{pos}}(x, y)& = &\rho(||x - y||),\quad x,y\in \R^3,  \\
        k_{\textit{or}}(v,w)&=&\gamma(v . w),\quad v,w\in \mathbb{S}^2,
    \end{array}
    \label{eq: kernel}
\end{equation*}
then for any two varifolds $\mu,\nu$, and any rigid motion $\phi$ on $\R^3$, we have $$\langle \phi_\#\mu, \phi_\#\nu \rangle_V=\langle \mu, \nu\rangle_V.$$
\end{theorem}
This result means that given a rigid motion $\phi$, $\langle\phi(M),\phi(N)\rangle_V =\langle M, N\rangle_V$.

The kernel $k_{\textit{pos}}$ is usually chosen as the Gaussian kernel $k_{\textit{pos}} = e^{-\frac{||x-y||^2}{\sigma^2}}$, with the scale parameter $\sigma$ needed to be tuned for each application.

\par
Kaltenmark~\etal~\cite{kaltenmark2017general} proposed several function for the $\gamma$ function of the spherical kernel. In this paper we retained the following functions: $\gamma(u)=u$ -- \textit{currents}, $\gamma(u)=e^{2u / \sigma_o^2}$ -- \textit{oriented varifolds}, and we propose  $\gamma(u) = |u|$ -- \textit{absolute varifolds}. For such kernels, two surface varifolds $M,N$ with ``similar" support (for example, if $M$ is a reparametrization of $N$, or if they represent two human shapes with the same pose but different body types) will have relatively small distance in the space of varifolds, so that $\langle M,N\rangle_V^2\simeq \langle M,M\rangle_V\langle N,N\rangle_V$, that is, they are almost co-linear. On the other hand, surface varifolds with very distant support will be almost orthogonal ($\langle M,N\rangle_V\simeq 0$) because of the Gaussian term in $k_{pos}$. Obviously, shapes that have some parts that almost overlap while others are far away will be in-between. Combined with its rotational invariance, this leads us to believe that the kernel product can be used to differentiate between poses and motions independently of body types.

\subsection{Comparing 3D Human Sequences}
\begin{figure}
    \centering
    \includegraphics[width=0.7\linewidth]{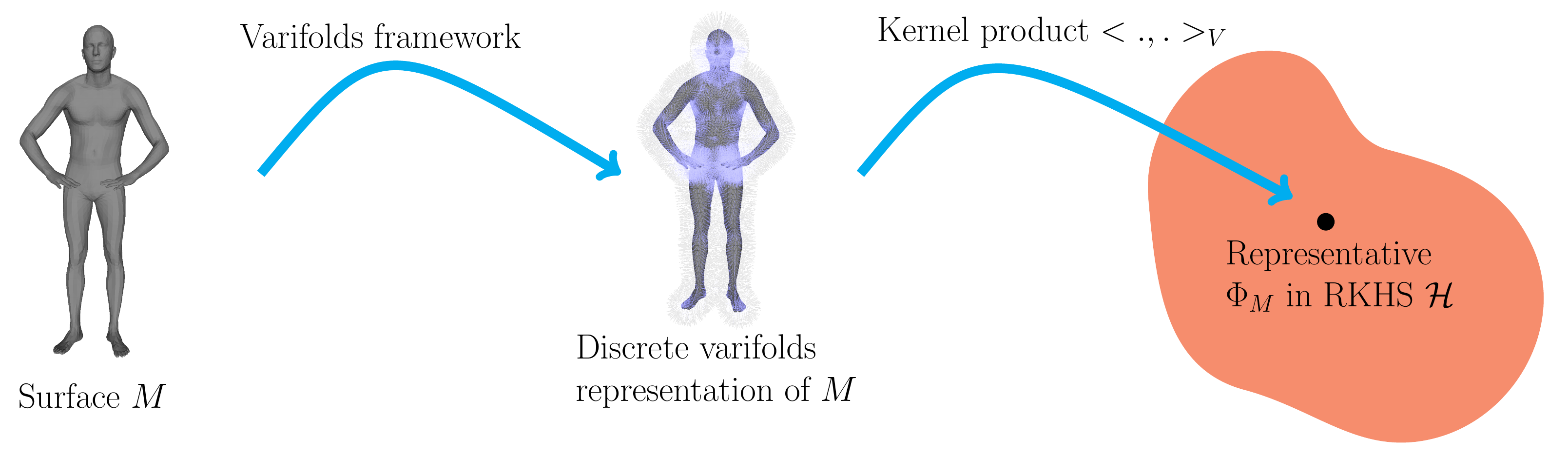}
    \caption{An overview of varifolds framework. First, a mesh M is transformed into its corresponding varifold representation. Then, the kernel product defined in~\Cref{eq: kernel_full}, transforms it into a representant $\Phi_M$ living in the Hilbert space $\mathcal{H}$ of this varifold.}
    \label{fig:summary_comparing}
\end{figure}

We need a way to compare sequences of $3D$ shapes $M_1, ..., M_T$, with $T$ possibly differing between sequences. For this, we use the kernel product $\langle ., .\rangle_V$ as a similarity metric. Thanks to the reproducing property of positive definite kernels~\cite{aronszajn1950theory}, it defines a reproducing kernel Hilbert space $\mathcal{H}$ (RKHS) which is an (infinite dimensional) Euclidean space endowed with an inner product corresponding to the kernel product, as described in the previous section. Any shape $M$ has a corresponding representative $\Phi_M:\R^3\times\mathbb{S}^2\rightarrow\R$ in this space, such that $\langle \Phi_M, \Phi_N \rangle_V = \langle M, N \rangle_V$ (\Cref{fig:summary_comparing}).
\paragraph{Modeling dynamics of temporal sequences.}
Thanks to the varifolds representation and the kernel product $\langle ., . \rangle$, the temporal sequence $M_1, ..., M_t$ corresponding to a motion in the human shape space can be seen as a temporal sequence $\Phi_{M_1}, ..., \Phi_{M_T}$ in the RKHS $\mathcal{H}$. Plus, since the varifold kernel is equivariant to rigid transformations, the product of two shapes within a sequence is invariant to any rigid transformation \textit{applied to the full motion}.  The Gram matrix $J_{ij} = \langle \Phi_{M_i}, \Phi_{M_j}\rangle $, which is a rigid transformation invariant matrix, would be a natural representant of the motion. However, its size vary with the length $T$ of the sequence. Inspired by Auto-Regressive (AR) models of complexity $k$ defined by  $\Phi_{M_t} = \sum_{i=1}^{k} \alpha_i \Phi_{M_{t-i}}$, several representations~\cite{slama2015accurate,turaga2008statistical} have been proposed for dynamical systems modeling. Hankel matrices~\cite{li2012cross} are one of the possible representations. The Hankel matrix of size $r, s$ corresponding to our time series $\Phi_{M_1}, ..., \Phi_{M_T}$ is defined as:
\begin{equation}
    \mathbf{H}_t^{r, s} = \left(\begin{array}{ccccc}
    \Phi_{M_1} & \Phi_{M_2} & \Phi_{M_3} & ... & \Phi_{M_s}  \\
    \Phi_{M_2} & \Phi_{M_3} & \Phi_{M_4} & ... & \Phi_{M_{s+1}}  \\
    ... & ... & ... & ... & ...  \\
    \Phi_{M_r} & \Phi_{M_{r+1}} & \Phi_{M_{r+2}} & ... & \Phi_{M_{r+s}} \\
    \end{array}\right)
    \label{eq:hankel_def}
\end{equation}
The rank of such matrix is usually, under certain conditions, the complexity $k$ of the dynamical system of the sequence. The comparison of two time series therefore become a comparison of high dimensional matrices. 
\par
\begin{figure}
    \centering
    \includegraphics[width=0.6\linewidth]{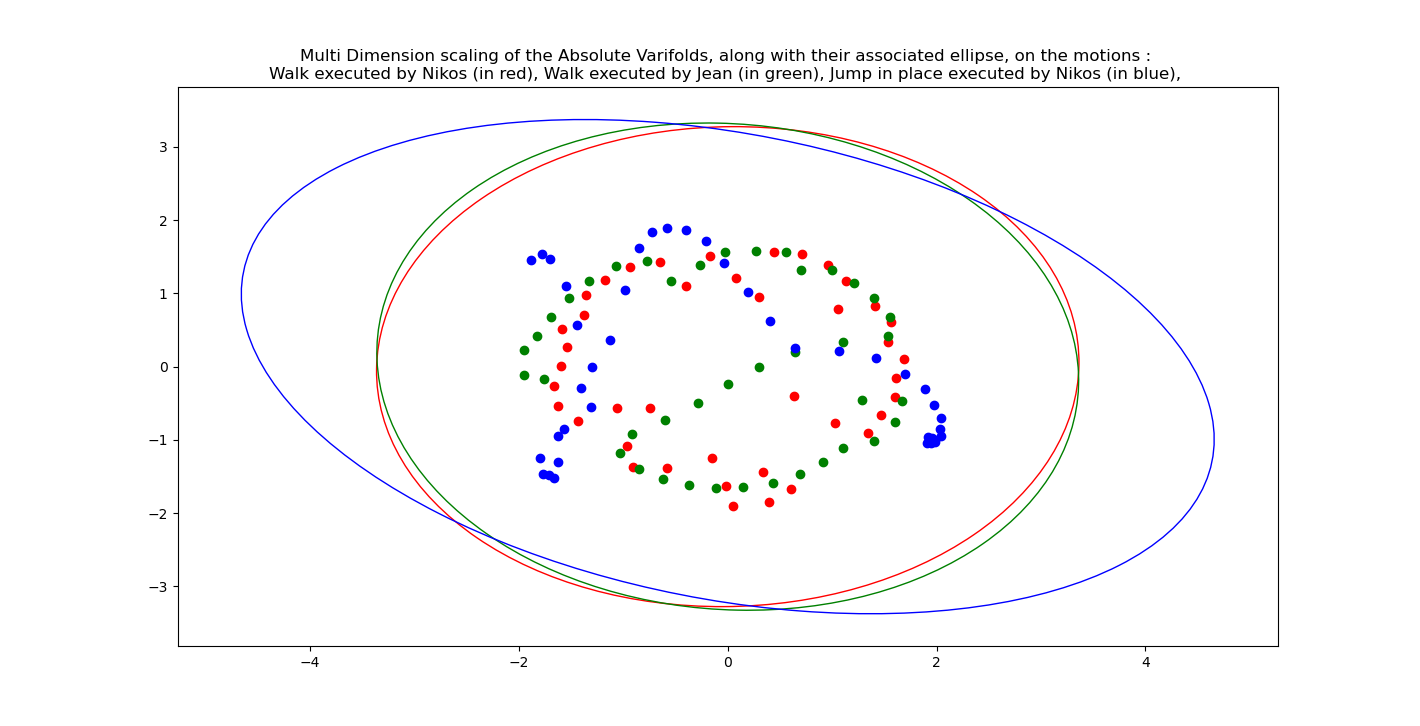}
    \caption{MDS illustration of three motions of CVSSP3D dataset, along with ellipse associated to their covariance.}
    \label{fig:interpretation}
\end{figure}

 
It is not straightforward to use those matrices since our shape representatives live in infinite dimensional space. A first idea would be to think about the Nystrom reduction method~\cite{williams2001using} to build an explicit finite dimensional representation for $\Phi_M$, but this would involve intensive computations. 
Another possibility is to think about the Gram matrix $\mathbf{H}\mathbf{H}^T$ derived from the Hankel matrix $\mathbf{H}$~\cite{zhang_efficient_2016,li2012cross}. We cannot directly derive the same kind of matrices since our representatives live in an infinite dimensional space. The Gram matrix of the motion, $J$, however, preserves the linear relationships of the AR model. We therefore derive the following matrix:


\begin{definition}
The Gram-Hankel matrix of size $r$, $G \in M_r(\mathbb{R})$ of the sequence $\Phi_{M_1}, ..., \Phi_{M_T}$ is defined as:
\begin{equation}
\label{equ:GramHankel}
    \mathbf{G}_{ij} = \sum_{k=1}^{T-r} \langle \Phi_{M_{i+k}}, \Phi_{M_{j+k}}\rangle = \sum_{k=1}^{T-r} \langle M_{i+k},M_{j+k}\rangle_V
\end{equation}
\end{definition}
We normalize $\mathbf{G}$ relatively to the Frobenius norm, following recommended practices~\cite{zhang_efficient_2016}. This matrix is the sum of the diagonal blocks $B^r_l$of size $r$ of the Gram matrix of the sequence pairwise inner products. 
A possible way of interpreting what encodes a single block $B^r_l$ of size $r$ when $r \geq k$ is to follow the idea of~\cite{kacem2018novel} the polar decomposition of the coordinate matrix of $\Phi_{M_{l}}, ... \Phi_{M_{r+l}}$. This coordinate matrix exists in the space $span(\Phi_{M_0}, ... \Phi_{M_k})$ under to the AR model hypothesis (any $\Phi_{M_{j}}$ is a linear combination of the first $k$ $\Phi_{M_i}$), and can be factorized into the product $U_lR_l$, where $U_l$ is an orthonormal $r \times k$ matrix, and $R_l$ an SPD matrix of size $k$. The matrix $R_l$ is the covariance (multiplied by $r^2$) of $\Phi_{M_{l}}, ... \Phi_{M_{r+l}}$ in $span(\Phi_{M_0}, ... \Phi_{M_k})$, and it encodes in some way its shape in this space. An illustration of such encoding is given in~\Cref{fig:interpretation}. For three motions from CVSSP3D dataset, we compute the varifold distances~\Cref{eq: kernel_full} between all samples of the motion. We then used Multidimensional Scaling (MDS)~\cite{cox2008multidimensional} to visualize them in a 2D space. We display the ellipse associated to the covariance of each motion. We see that the one associated to jump in place (blue) motion is distinguishable from the ones associated to walk motions (red and green).

The Gram matrix block $B_l^r$ is written as $B_l^r = U_l R_l^2 U_l^T$ and contains such information. Searching for the complexity $k$ of the AR model would be sensitive to errors, and computing the associated $R_l$ for comparisons with an SPD metric would be time consuming in our case. We thus preferred the rather simpler Gram-Hankel matrix, that cancels possible noise in single blocks when summing them. Finally, using the Frobenius distance $d(\mathbf{G_i}, \mathbf{G_j}) = ||\mathbf{G_i} - \mathbf{G_j}||_F$ where $G_i$ and $G_j$ are two Gram-Hankel matrices, to compare two motions lead us to rather good results. The blocks of size $r$ are expressive enough when $r \geq k$, and taking their sum will ensure us to cancel possible noise added to a single block. The degenerate nature of $G$, does not allow for efficient use of SPD metrics such as the Log-Euclidean Riemannian Metric (LERM) on the $G_i$ (more details are available in the supplementary material).
With this approach, the comparison of two human motions is formulated as the comparison of two symmetric positive semi-definite matrices. 

\begin{proposition}
The Gram-Hankel matrix $G$ associated to a motion $M_1, ..., M_T$ defined by~\Cref{equ:GramHankel} has the following properties:
\begin{enumerate}
    \item It is invariant to parameterization  (property of the kernel product).
    \item It is invariant to rigid transformation applied to a motion.
\end{enumerate}
\end{proposition}

 \paragraph{Normalizations}
As the definition of the kernel shows the method is not invariant to scale, we normalize the inner products as following: $\left\langle\frac{M_i}{||M_i||_V},\frac{M_j}{||M_j||_V}\right\rangle_V.$

\par 
While our method is translation invariant, the use of the Gaussian kernel implies that the product will be near 0 when the human shapes are at long range. To avoid this, we translate the surface $M$ with triangles $T_1, ..., T_m$ by its centroid $c_M=  \frac{\sum_{i=1}^m a_ic_i}{\sum_{i=1}^m a_i}$,  where $c_i$ and $a_i$ correspond to the center and area of triangle $T_i$. We apply $M \mapsto M - c_M$ before computing the products.



\section{Experiments}
Computing varifold kernel products can often be time consuming, due to the quadratic cost in memory and time in terms of vertex number for computing $\langle M, N \rangle_V$. However, the recent library Keops~\cite{JMLR:v22:20-275}, designed specifically for kernel operations proposes efficient implementations with no memory overflow, reducing time computation by two orders of magnitudes. We used those implementations with the Pytorch backend on a computer setup with Intel(R) Xeon(R) Bronze 3204 CPU @ 1.90GHz, and a Nvidia Quadro RTX 4000 8GB GPU.
\subsection{Evaluation setup}
In order to measure the performance in motion retrieval, we use the classical performance metrics used in retrieval: Nearest neighbor (NN), First-tier (FT) and Second-tier (ST) criteria. 
For each experiment, we take $r$ values ranging from $1$ to $T_{min}$ where $T_{min}$ is the minimal sequence length in the dataset. We also take 10 $\sigma$ values for the Gaussian kernel ranging from $0.001$ to $10$ in log scale. The score displayed is the best score among all $r$ and $\sigma$ values. For oriented varifolds, the $\sigma_o$ of the gamma function is fixed to $0.5$ as in~\cite{kaltenmark2017general}.

\subsection{Datasets}
\begin{figure}
        \centering
        \begin{subfigure}[b]{0.33\linewidth}
            \includegraphics[width=\linewidth]{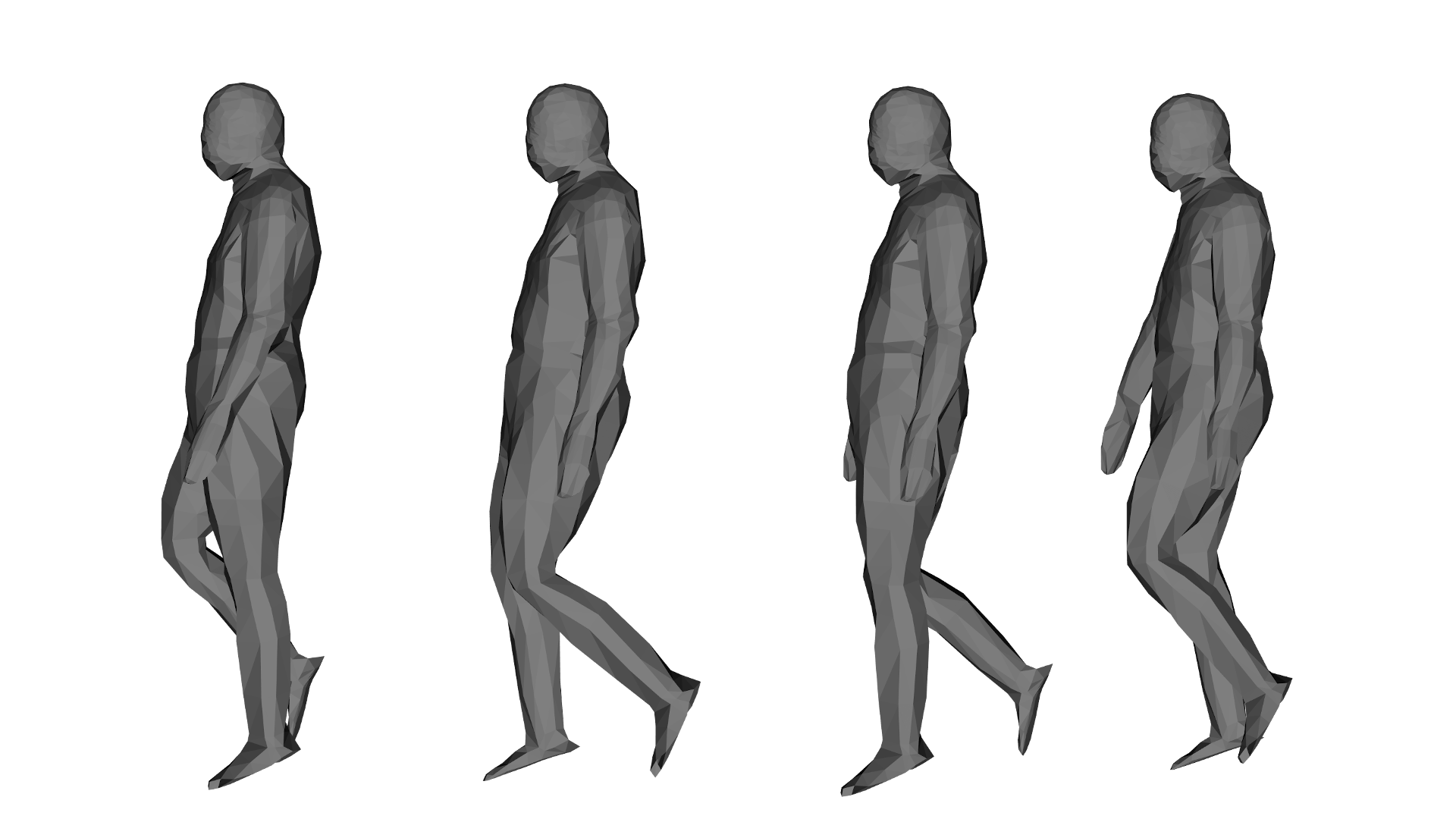}
            \caption{}
        \end{subfigure}
        \begin{subfigure}[b]{0.28\linewidth}
            \includegraphics[width=\linewidth]{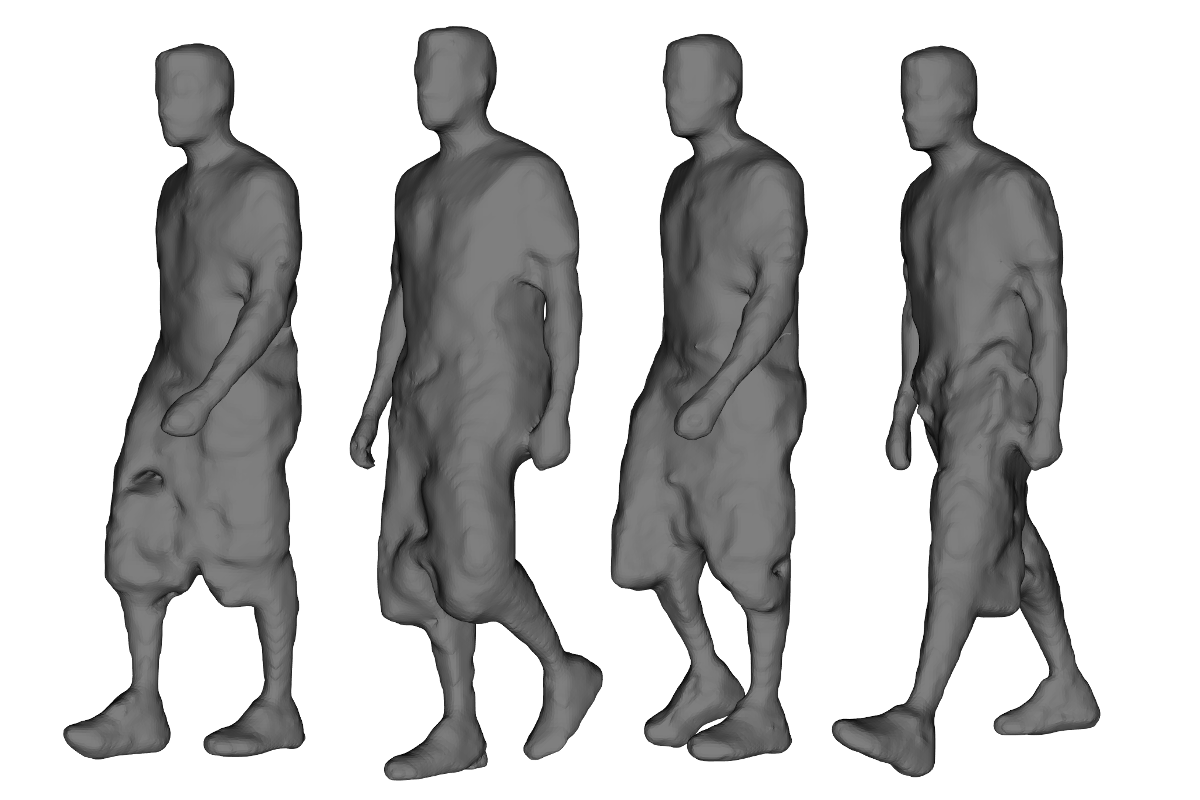}
            \caption{}
        \end{subfigure}
        \begin{subfigure}[b]{0.3\linewidth}
            \includegraphics[width=\linewidth]{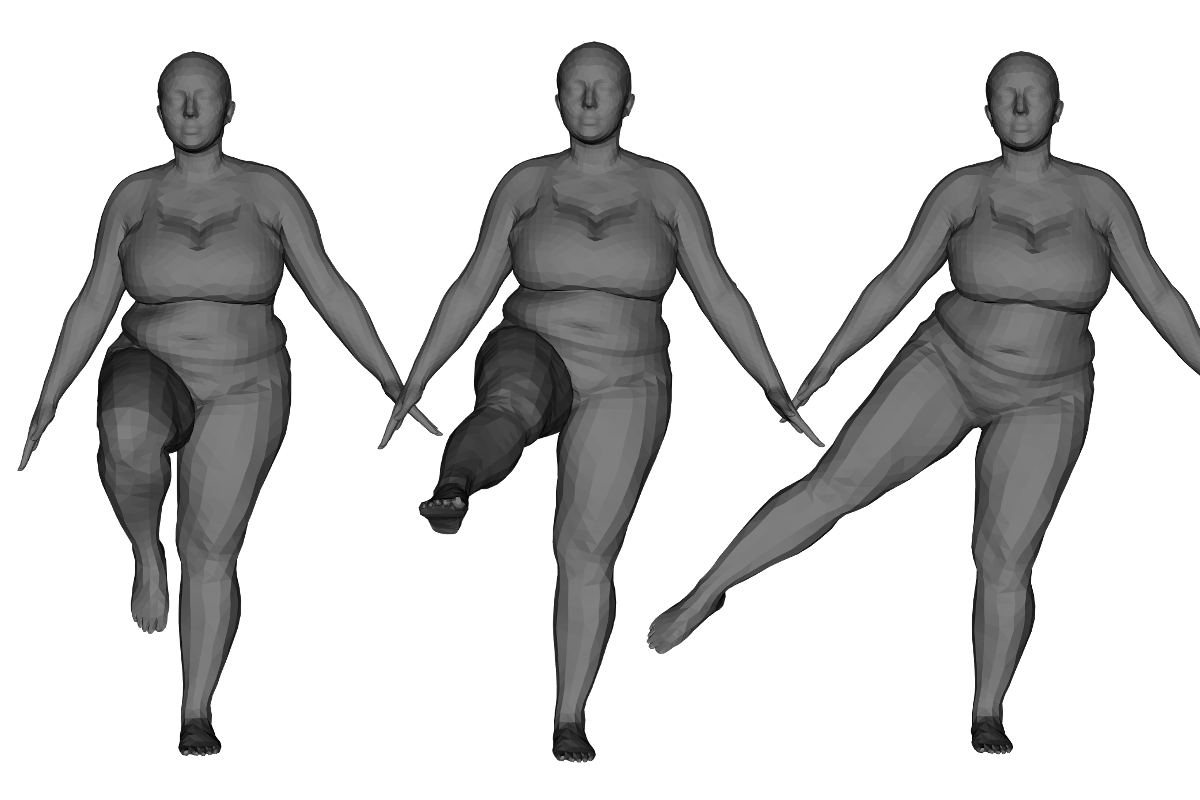}
            \caption{}
        \end{subfigure}
        \caption{Example motions from the datasets: (a) Slow walking motion from CVSSP3D synthetic dataset, (b) Walking motion from the CVSSP3D real dataset, (c) Knees motion from the Dyna dataset.}
        \label{fig:synth_data}
    \end{figure}

\textbf{CVSSP3D synthetic dataset}~\cite{cvssp3d}. A synthetic model (1290 vertices and 2108 faces) is animated thanks to real motion skeleton data. Fourteen individuals executed 28 different motions: sneak, walk (slow, fast, turn left/right, circle left/right, cool, cowboy, elderly, tired, macho, march, mickey, sexy, dainty), run (slow, fast, turn right/left, circle left/right), sprint, vogue, faint, rock n’roll, shoot.  An example of human motion from this dataset is presented in~\Cref{fig:synth_data}. The frequency of samples is set to 25Hz, with 100 samples per sequences. The maximum computation time of Gram-Hankel matrix is 0.89s.\\
\textbf{CVSSP3D real dataset}~\cite{gkalelis2009i3dpost}. This dataset contains reconstructions of multi view performances. 8 individuals performed 12 different motions: walk, run, jump, bend, hand wave (interaction between two models), jump in place, sit and stand up, run and fall, walk and sit, run then jump and walk, handshake (interaction between two models), pull. The number of vertices vary between 35000 and 70000. The frequency of samples is also set to 25Hz, and sequence length vary from 50 to 150 (average 109). We keep the 10 individual motions following~\cite{veinidis2019effective}. An example motion of the dataset is displayed in~\Cref{fig:synth_data}(b). The maximum computation time of Gram-Hankel matrix is 6m30s.
    The sensitivity of the reconstruction pipeline to the clothes is illustrated by the presence of noise as illustrated in ~\Cref{fig:challenges}. This noise makes this dataset challenging for 3D shape human shape comparison and for 3D human motion retrieval.
    
\textbf{Dyna dataset~\cite{pons2015dyna}}. This dataset is created from 4D human scans. A human template (6890 vertices) is registered to human body scans sampled at 60 Hz, and sequence length vary from 150 to 1200 (average 323). 10 individuals performed at most 14 different treadmill motions (hips, knees, light hopping stiff, light hopping loose, jiggle on toes, one leg loose, shake arms, chicken wings, punching, shake shoulders, shake hips, jumping jacks, one leg jump, running on spot), which means that the individual only move along the height axis. 
An example of human motion from Dyna dataset is presented in~\Cref{fig:synth_data}(c). The maximum computation time of Gram-Hankel matrix is 2m30s.


\subsection{Motion Retrieval on CVSSP3D Dataset}

\textbf{Comparison with state-of-the-art.} 
We compare our motion retrieval approach to the best features presented in~\cite{veinidis2019effective,pierson2021projection} and deep learning descriptors:
(1) The 3D harmonics descriptor~\cite{papadakis20083d}\cite{veinidis2019effective} is a descriptor based on point cloud repartition in space,
(2) Breadths spectrum Q-breadths and Q-shape invariant~\cite{pierson2021projection} are presented as 2 fully invariant descriptors derived from convex shape analysis, 
(3) Aumentado-Armstrong \etal~\cite{gdvae_2019}~propose a human pose latent vector in their Geometrically Disentangled Variational AutoEncoder (GDVAE), 
(4) Zhou~\etal~\cite{zhou20unsupervised} propose a human pose latent vector derived from the Neural3DMM~\cite{neural3dmm_2019} mesh autoencoder architecture, 
and (5) Cosmo \etal~\cite{cosmo2020limp} propose a human pose latent vector in a similar approach as GDVAE, called Latent Interpolation with Metric Priors (LIMP). For the artificial dataset, the optimal $\sigma$ were fixed to $0.17$ for current, $0.17$ for absolute varifolds, and $0.02$ for oriented varifolds. The optimal $r$ were $97.92$ and $90$ for current, absolute and oriented varifolds respectively.
\begin{table*}[!ht]
\centering
\scalebox{0.95}{\begin{tabular}{l|cc|ccc|ccc|ccc}
 \multirow{2}{*}{Representation} & \multirow{2}{*}{$\Gamma$ inv.} & \multirow{2}{*}{$SO(3)$} &\multicolumn{3}{l|}{Artificial dataset} & \multicolumn{3}{l|}{Real dataset} & \multicolumn{3}{l}{Dyna dataset} \\ \cline{4-12}
 & & & NN & FT & ST & NN & FT & ST & NN & FT & ST\\ \hline 
Shape Dist.~\cite{OsadaACM2002}\cite{veinidis2019effective} & \cmark & \cmark & 92.1 & 88.9 & 97.2 &  77.5 & 51.6 & 65.5  & / & / & / \\
Spin Images~\cite{JohnsonPAMI99}\cite{veinidis2019effective} & \cmark & \cmark & \textbf{100} & 87.1 & 94.1 &  77.5 & 51.6 & 65.5  & / & / & / \\
3D harmonics~\cite{veinidis2019effective} & $\approx$ & $\approx$ & \textbf{100} & 98.3 & 99.9 &  92.5 & \textbf{72.7} & \textbf{86.1}  &  / & / & /\\
Breadths spectrum~\cite{pierson2021projection} & \cmark & \cmark & \textbf{100} & 99.8 & \textbf{100} & / & / & / & / & / & / \\
Shape invariant~\cite{pierson2021projection} & \cmark & \cmark & 82.1 & 56.8 & 68.5 & / & / & / & / & / & / \\
Q-Breadths spectrum~\cite{pierson2021projection} & $\approx$ & \cmark & / & / & / & 80.0 & 44.8 & 59.5 & / & / & /  \\
Q-shape invariant~\cite{pierson2021projection} & $\approx$ & \cmark & / & / & / & 82.5 & 51.3 & 68.8 & / & / & /  \\
Areas~\cite{pierson2021projection} & \cmark & \xmark & / & / & / & / & / & / & 37.2 & 24.5 & 35.8 \\
Breadths~\cite{pierson2021projection} & \cmark & \xmark & / & / & / & / & / & / & 50.7 & 36.2 & 50.5 \\
Areas \& Breadths~\cite{pierson2021projection} & \cmark & \xmark & / & / & / & / & / & / & 50.7 & 37.2 & 51.7 \\ 
GDVAE~\cite{gdvae_2019}& \cmark & \cmark & \textbf{100} & 97.6 & 98.8 & 38.7 & 31.6 & 51.6 & 18.7 & 19.6 & 32.2 \\
Zhou \etal~\cite{zhou20unsupervised} &  \xmark & \xmark & \textbf{100} & 99.6 & 99.6 & / & / & / &  50.0 & 40.4 & 57.0  \\
LIMP~\cite{cosmo2020limp} & \cmark & \xmark & \textbf{100} & \textit{99.98} & \textit{99.98}  & / & / & / &  29.1 & 20.7 & 33.9 \\ 
\textit{SMPL pose vector}~\cite{SMPL:2015} & $\approx$ & \cmark & / & / & / & / & / & / & \textit{58.2} & \textbf{\textit{45.7}} & \textbf{\textit{63.2}} \\ \hline \hline 
Current & \cmark & \cmark & \textbf{100} & \textbf{100} & \textbf{100} &   92.5 & 66.0 & 78.5 & 59.0 & 34.1 & 50.4 \\
Absolute varifolds & \cmark & \cmark & \textbf{100} & \textbf{100} & \textbf{100} & \textbf{95.0} & 66.6 & 80.7 & \textbf{60.4} & 40.0 & 55.9  \\
Oriented varifolds & \cmark & \cmark & \textbf{100} & \textbf{100} & \textbf{100} &  93.8 & 65.4 & 78.2  &  \textbf{60.4} & 40.8 & 55.9 \\
\end{tabular}}
\caption{Full comparison of motion retrieval approaches. First two columns correspond to group invariance ($\Gamma$: reparameterization group, $SO(3)$: rotation group), telling whether or not the required invariance is fullfilled (\cmark: fully invariant, $\mathbf{\approx}$ : approximately invariant (normalization, supplementary information, ...), \xmark: no invariance). Remaining columns correspond to retrieval scores, where the '/' symbol means that there is no result for the method on the given dataset for various reasons, such as unavailable implementation or the method not being adapted for the dataset (for example, in line 470, [41] is based on a given mesh with vertex correspondences  and cannot  be applied to CVSSP3D real dataset). The results are displayed for CVSSP3D artificial and real datasets, and Dyna datasets. Our method is competitive or better than the approach consisting of combing DTW with any descriptor, while showing all required invariances. }
\label{tab:compar_tot}
\end{table*}

We observe the results on the CVSSP3D artificial dataset in~\Cref{tab:compar_tot}. Only our approach are able to get $100\%$ in all performance metrics. We also observe that it is the only approach able to outperform the LIMP learned approach.
\par 
For the real dataset, the optimal $\sigma$ were fixed to $0.06$ for current, $0.17$ for absolute varifolds and oriented varifolds. The optimal $r$ were $48, 43$ and $46$ for current, absolute and oriented varifolds respectively.

We observe the results on the CVSSP3D real dataset in~\Cref{tab:compar_tot}. Absolute varifolds approach outperforms by $2.5\%$ the 3D descriptor in terms of NN metric, while being less good for FT and ST. In terms of fully invariant methods, we outperform by $10\%$ the proposed approaches. The absolute varifolds methods is the best of our approach, but we do not observe significant sensitivity between different varifolds.
We finally observe that the point cloud descriptors of GDVAE has the lowest performance.
\subsection{Motion Retrieval on Dyna Dataset}
\par 
\textbf{Comparison with state-of-the-art.}
No benchmark exists on this dataset, a little has been made on the registrations provided by Dyna. We applied the following methods to extract pose descriptors and made pairwise sequences comparisons using dynamic time warping, in a similar protocol as~\cite{veinidis2019effective,pierson2021projection}, without the temporal filtering use for clothes datasets, since the dataset is not noisy. We compare our approach to descriptor sequences of the following approaches: (1) Areas and Breadths~\cite{pierson2021projection} are parameterization and translation invariants derived from convex shapes analysis, (2) The pretrained GDVAE on SURREAL is applied directly on the dataset, (3) the pretrained LIMP VAE on FAUST is applied directly on the dataset, (4) Zhou~\etal~\cite{zhou20unsupervised} provide pretrained weights on the AMASS dataset~\cite{amass} for their approach. This dataset shares the same human body parameterization as Dyna, so we can use the pretrained network on Dyna, and (5) The Skinned Multi-Person Linear model (SMPL) body model~\cite{SMPL:2015} is a parameterized human body model. We use the pose vector of the body model, computed in~\cite{bogo_dynamic_2017} using additional information.
For the Dyna dataset, the optimal $\sigma$ were fixed to $0.02$ for current $0.06$ for absolute varifolds, and $0.16$ for oriented varifolds. The optimal $r$ were $27, 31$ and $72$ for current, absolute and oriented varifolds respectively.

\begin{figure}[b]
    \centering
    \includegraphics[width=0.6\linewidth]{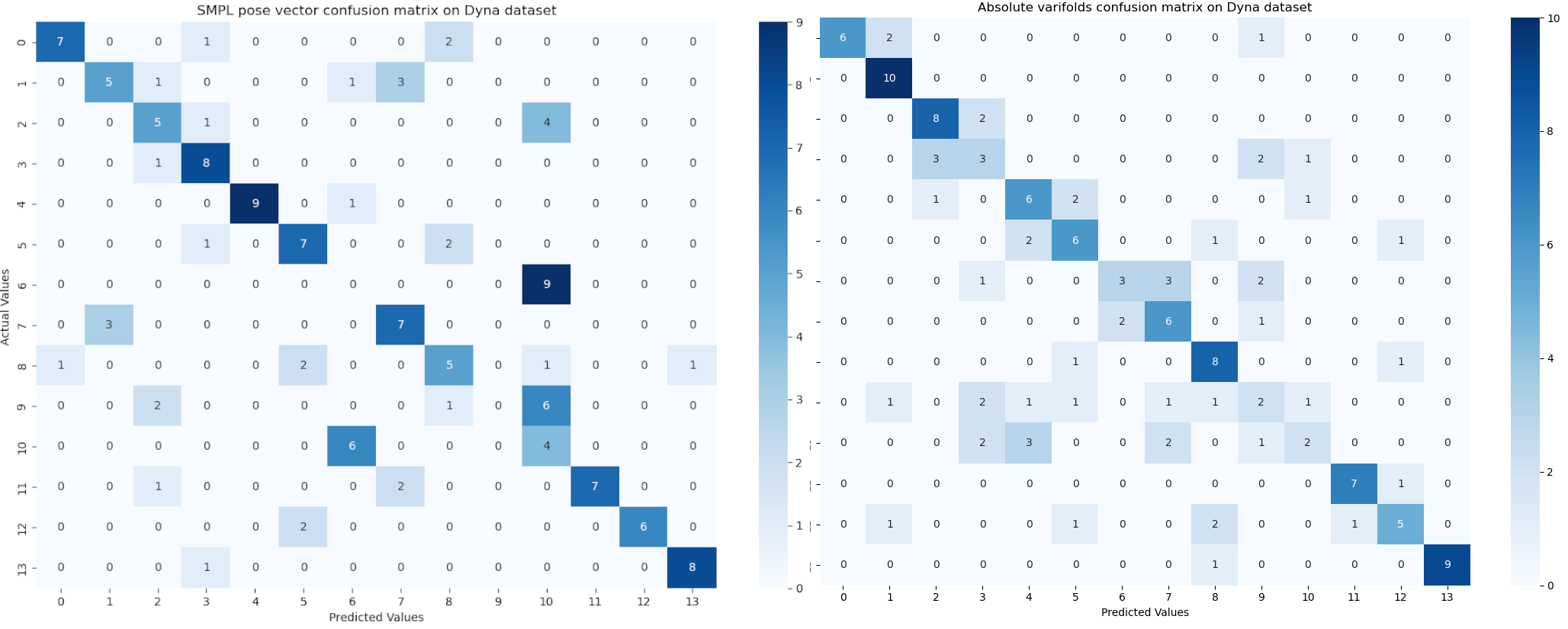}
    \caption{Confusion matrix of SMPL (left) and Oriented Varifolds (right) on the Dyna dataset.}
    \label{fig:confusion}
\end{figure}

As shown in~\Cref{tab:compar_tot} the oriented and absolute varifolds is the best by 2~\% in terms of NN metric compare to SMPL, and by more than 10~\% to other approaches, including the parameterization dependant approach of~\cite{zhou20unsupervised}. The FT and ST performance are however less good than SMPL. This can be explained by its human specific design, along with the costly fitting method, that use additional information (gender, texture videos). Finally, we observe that point cloud neural networks are not suitable for high set of complex motions.
\subsection{Qualitative analysis on Dyna dataset}

We display in~\Cref{fig:confusion} the Nearest Neighbor score confusion matrices for both SMPL and Oriented Varifolds. The confusion matrices for the other datasets are available in the supplementary material. We observe that on Dyna, the difficult cases were \textit{jiggle on toes, shake arms, shake hips and jumping jacks}, corresponding to l3, l6, l9 and l10 in confusion matrix. Our approach is able to classify better these motions than SMPL. In addition, SMPL was not able to retrieve as a Nearest Neighbor, a similar motion to shake arms or shake hips corresponding to l6 and l9. This Figure shows also that our approach retrieves perfectly the knees motion corresponding to l1. The~\Cref{fig:query} shows some qualitative results of our approach. It illustrates the first tier of a given query on Dyna dataset.
\begin{figure}[!ht]
    \centering
   \scriptsize
    \begin{subfigure}[b]{0.19\linewidth}
    
        \includegraphics[width=\linewidth]{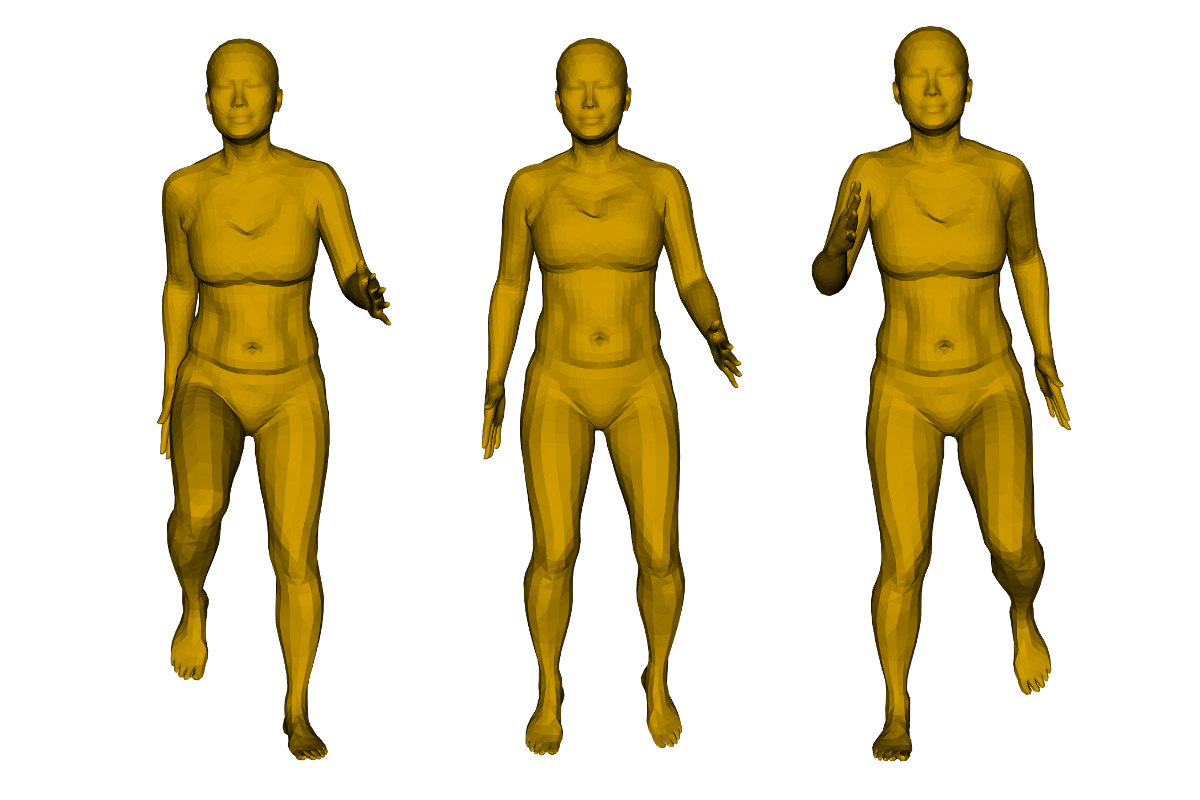}
        \caption{50020, running on spot}
    \end{subfigure}
    \begin{subfigure}[b]{0.19\linewidth}
        \includegraphics[width=\linewidth]{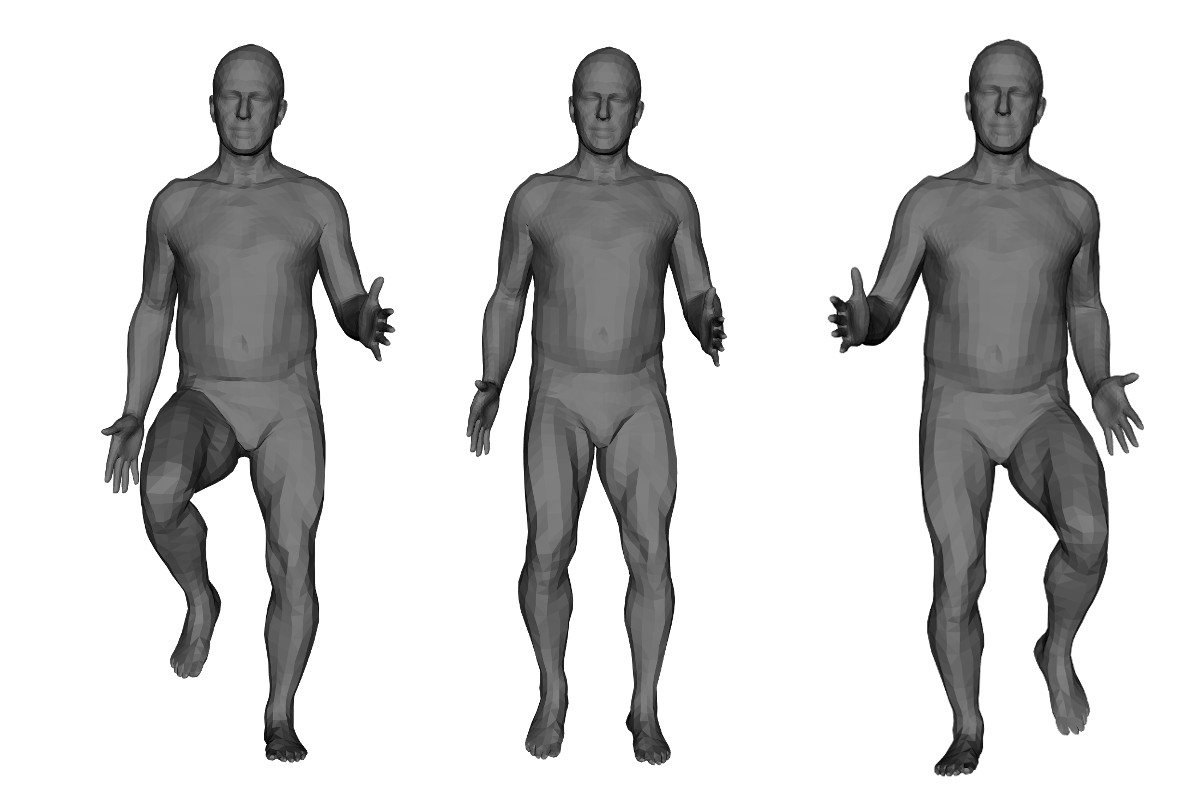}
        \caption{50026, running on spot}
    \end{subfigure}
    \begin{subfigure}[b]{0.19\linewidth}
        \includegraphics[width=\linewidth]{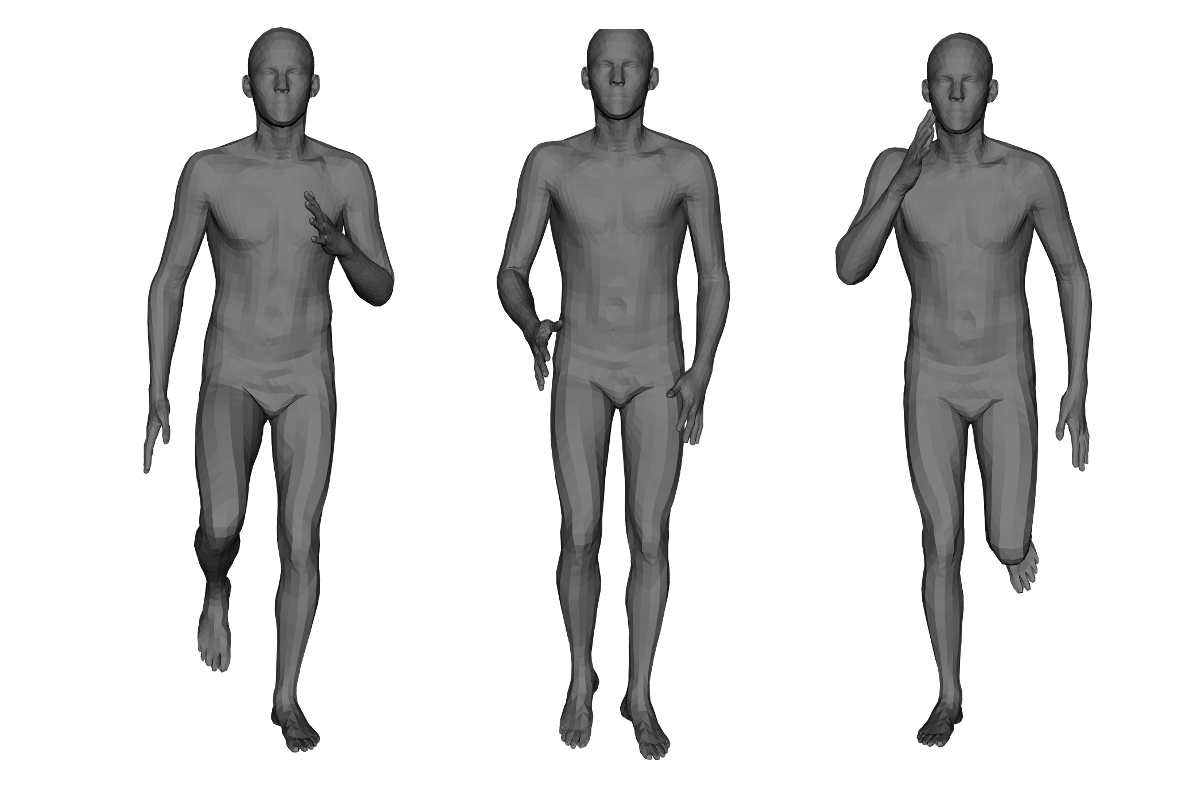}
        \caption{50009, running on spot}
    \end{subfigure}
    \begin{subfigure}[b]{0.19\linewidth}
        \includegraphics[width=\linewidth]{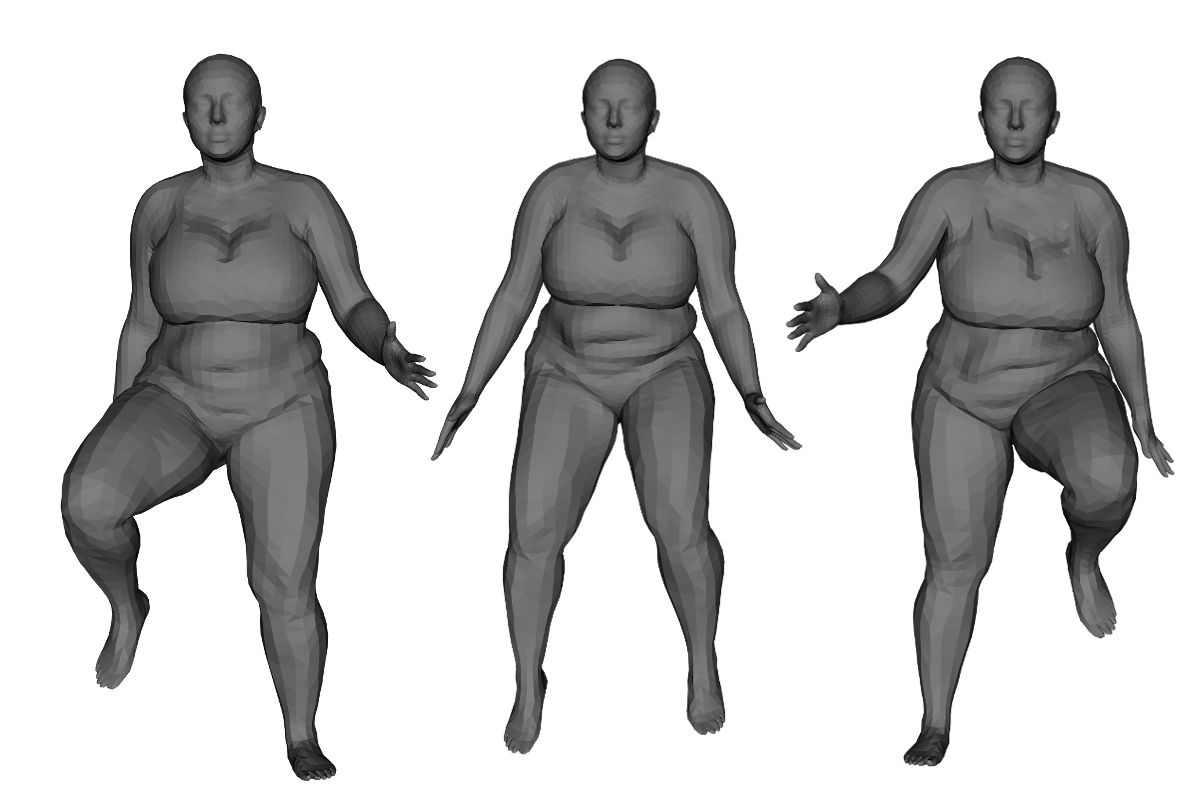}
        \caption{50004, running on spot}
    \end{subfigure}
    \begin{subfigure}[b]{0.19\linewidth}
        \includegraphics[width=\linewidth]{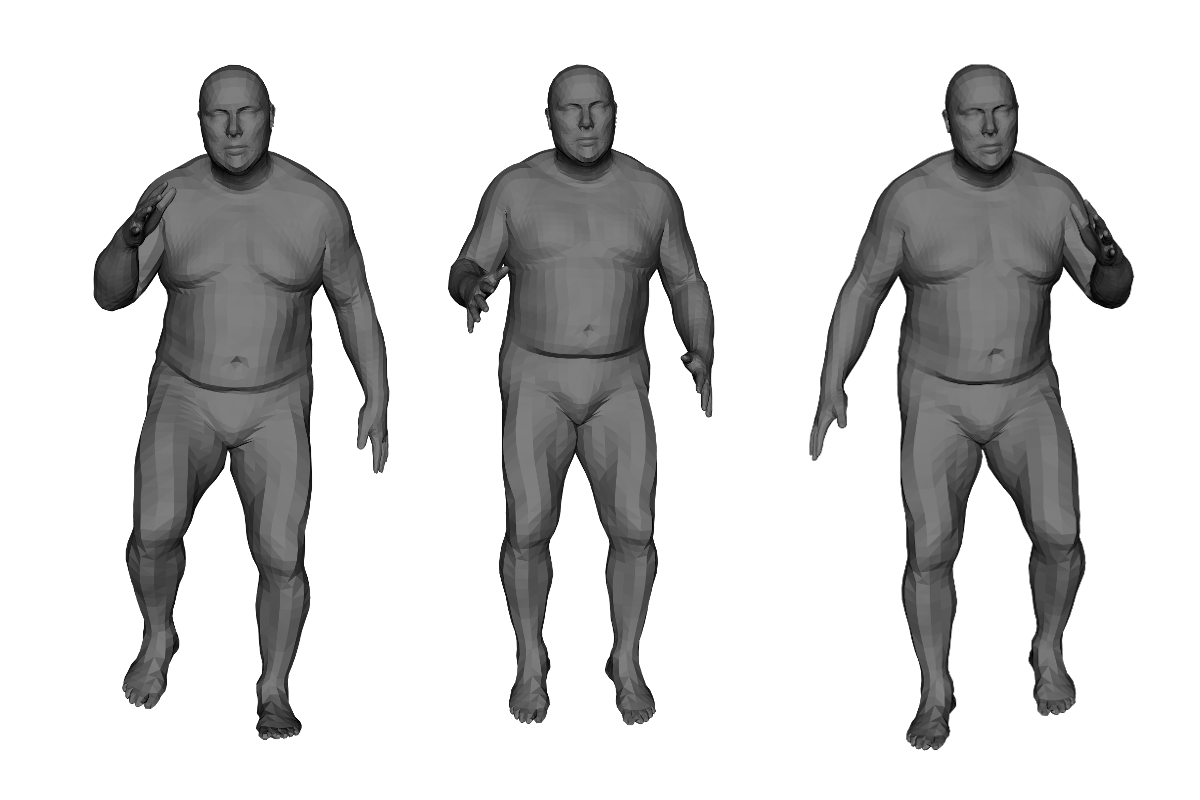}
        \caption{50007, running on spot}
    \end{subfigure}
    \begin{subfigure}[b]{0.19\linewidth}
        \includegraphics[width=\linewidth]{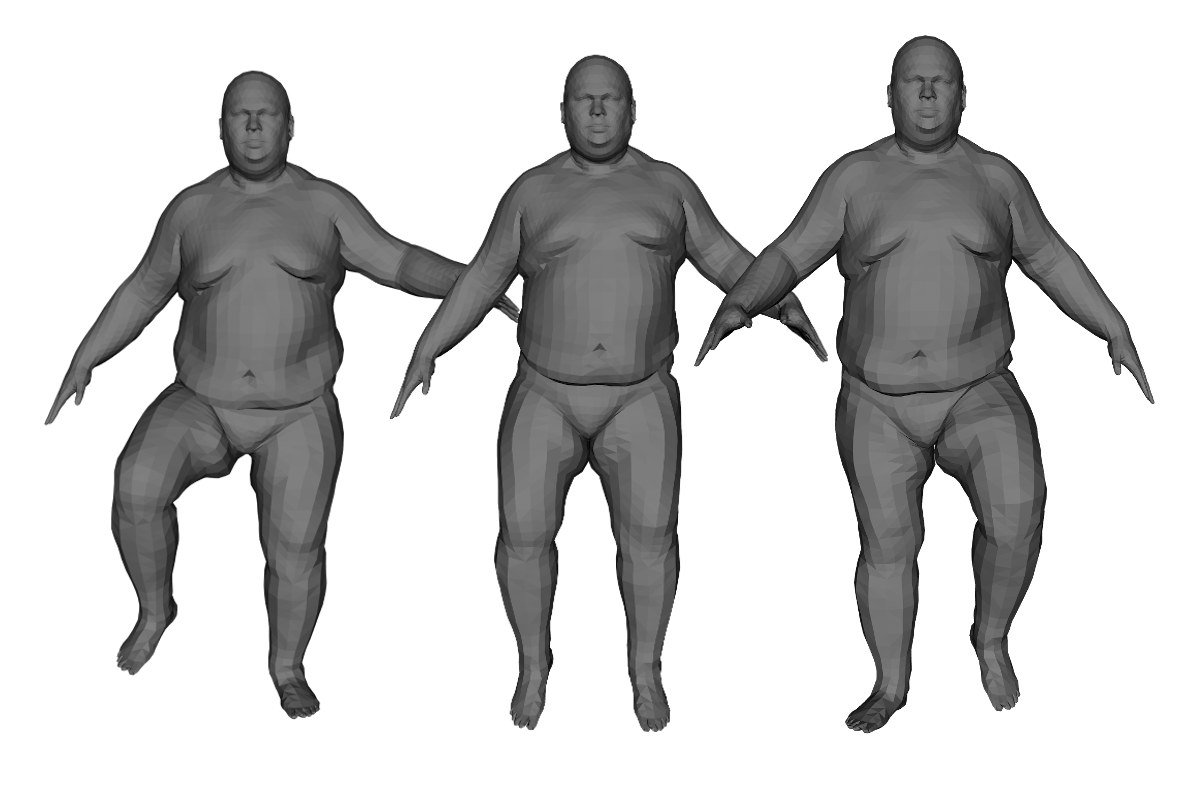}
        \caption{50002, running on spot}
    \end{subfigure}
    \begin{subfigure}[b]{0.19\linewidth}
        \includegraphics[width=\linewidth]{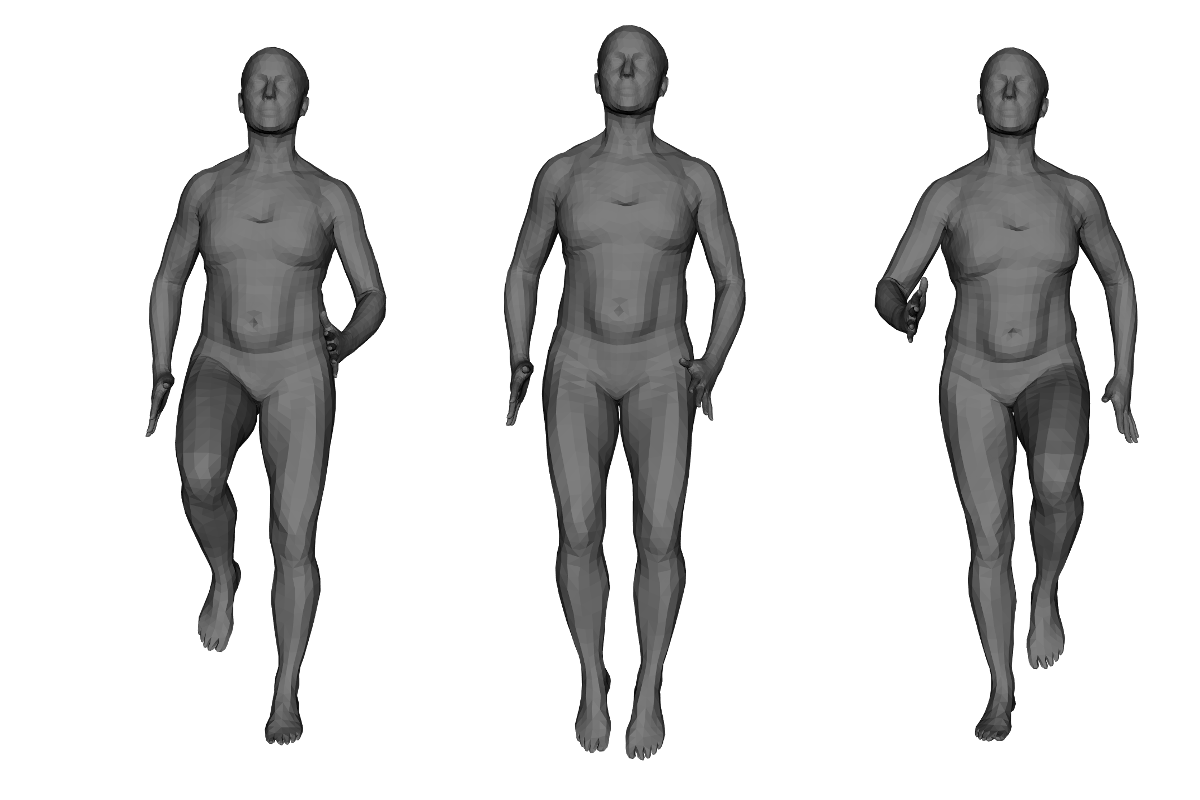}
        \caption{50025, running on spot}
    \end{subfigure}
    \begin{subfigure}[b]{0.19\linewidth}
        \includegraphics[width=\linewidth]{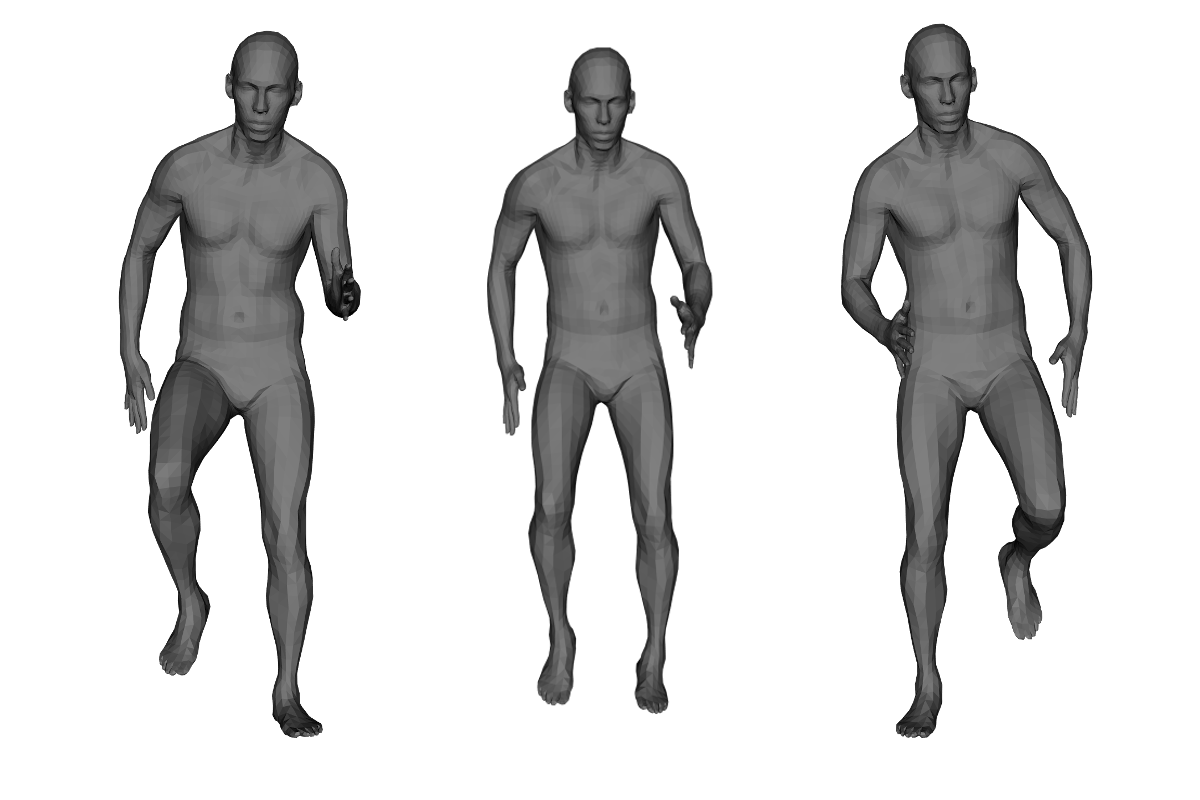}
        \caption{50027, running on spot}
    \end{subfigure}
    \begin{subfigure}[b]{0.19\linewidth}
        \includegraphics[width=\linewidth]{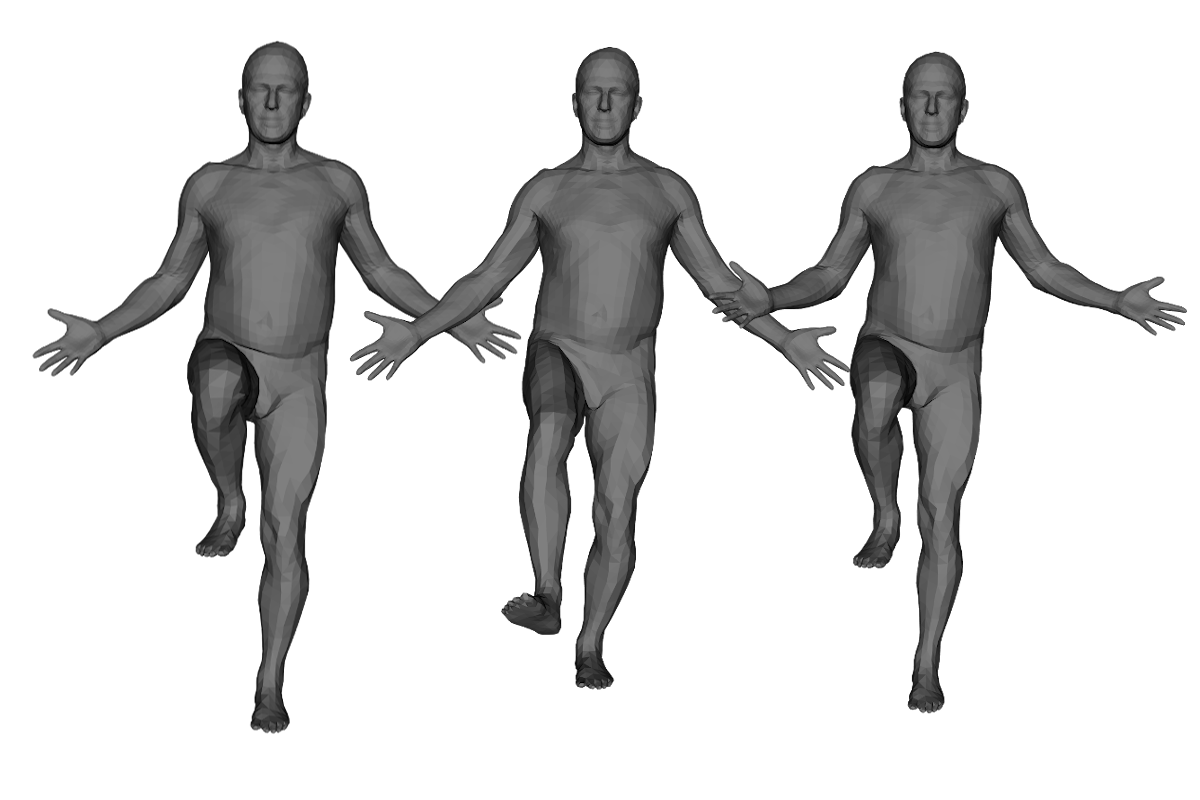}
        \caption{50026, one leg jump}
    \end{subfigure}
    \begin{subfigure}[b]{0.19\linewidth}
        \includegraphics[width=\linewidth]{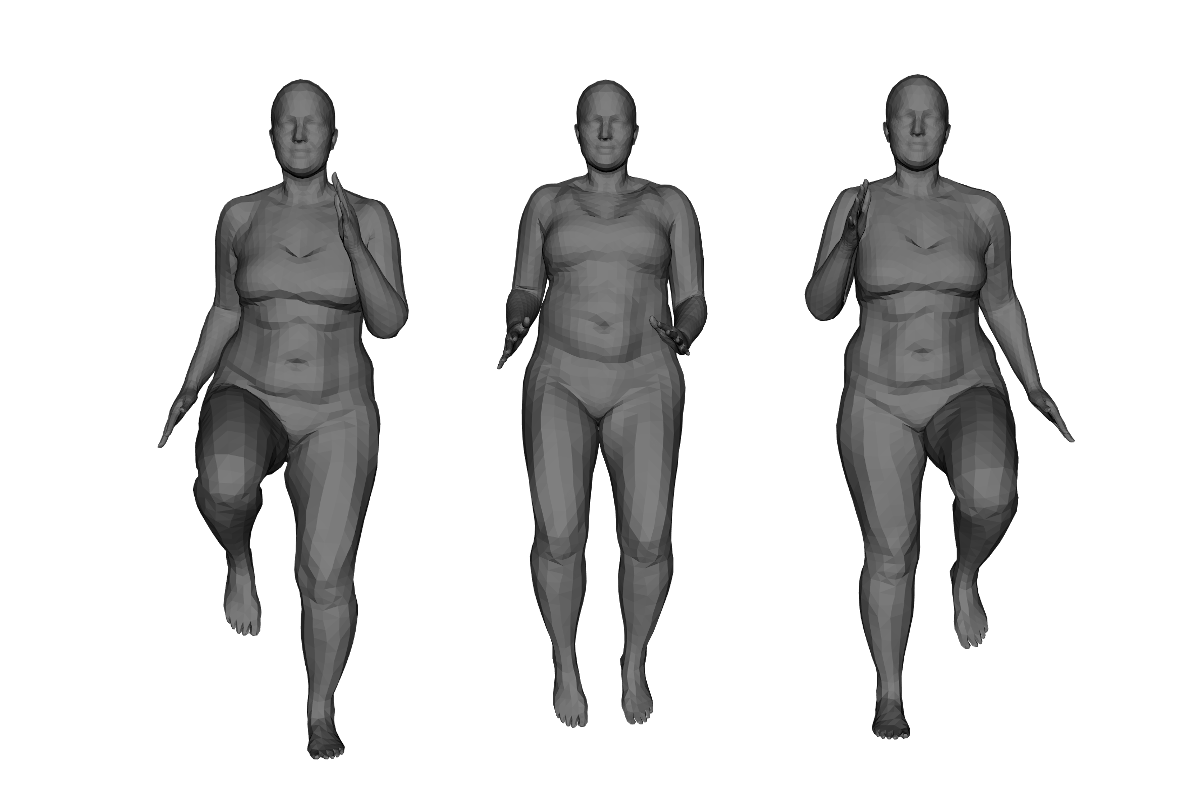}
        \caption{50022, running on spot}
    \end{subfigure}
    \caption{Qualitative results on Dyna dataset. Given the query corresponding to running on spot motion (a), the first tier result using oriented varifolds are given by (b) -- (j).}
    \label{fig:query}
\end{figure}

\section{Discussion}
\begin{figure}[!ht]
\begin{floatrow}
\ffigbox{
    \centering

    \includegraphics[width=0.9\linewidth]{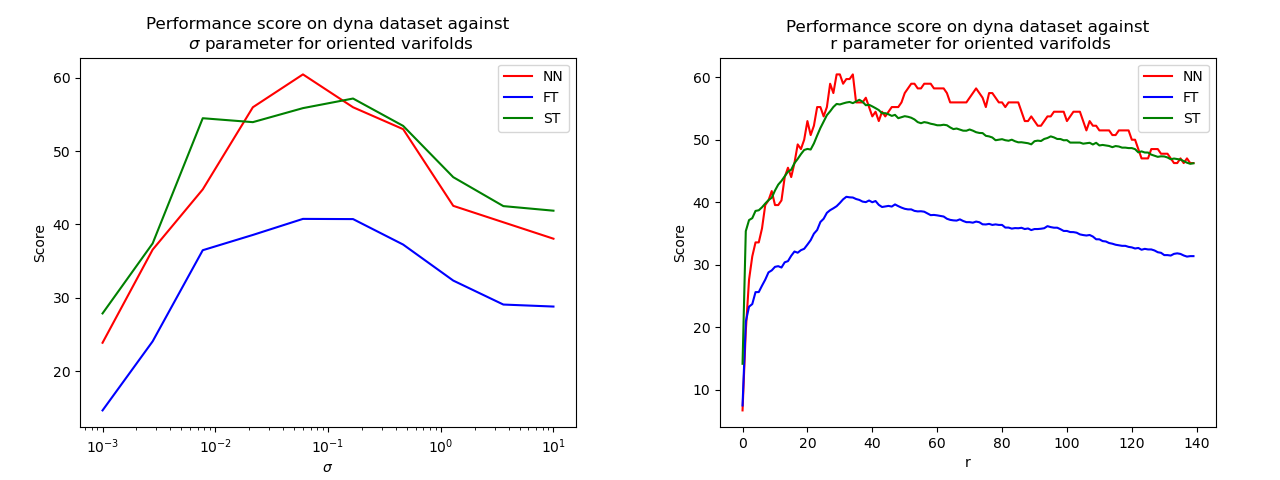}
    }{%
    \caption{NN, FT, ST metric relatively to the $\sigma$ parameters (left) and to the $r$ parameter (right) on Dyna dataset for oriented varifolds.}
    \label{fig:sigma_param}
}
\capbtabbox{%
    \centering
    \begin{tabular}{c|c|c|c|c}
        Centroid & Inner & NN & FT & ST \\
        \hline \hline 
        \xmark & \xmark & 51.5 & 34.6 & 53.4 \\
        \xmark & \cmark & 52.2 & 33.4 & 50.5 \\
        \cmark & \xmark & 59.7 & 40.7 & 55.8 \\ 
        \cmark & \cmark & \textbf{60.4} & \textbf{40.8} & \textbf{55.9} \\ \hline
    \end{tabular}
    }{%
    \caption{Retrieval performance of the normalizations on Dyna dataset, for oriented varifolds. Both are useful.}
    \label{tab:norm_or}
}
\end{floatrow}
\end{figure}

\paragraph{Effect of the parameters for oriented varifolds on Dyna dataset.}
We provide in~\Cref{fig:sigma_param}, the performance relative to the parameters $\sigma$ and $r$, for oriented varifolds on Dyna dataset. We observe that the choice of those parameters is crucial. We also display the performances of oriented varifolds with the 2 normalizations techniques, showing that they both help to obtain the best results. More discussion is provided in the supplementary material.

\paragraph{Limitations.}
Our approach presents two main limitations: (i) To measure distance between matrices, we have used Euclidean distance, which does not exploit the geometry of the symmetric positive semi-definite matrices manifold, (ii) There is no theoretical limitation to apply this framework to the comparison of other 3D shape sequences (eg. 3D facial expressions, or 3D cortical surfaces evolutions) other than that between body shape. However, in practice one should redefine the hyperparameters $(r,\sigma)$ of the Kernel (Theorem 1).

\section{Conclusion}
We presented a novel framework to perform comparison of 3D human shape sequences. 
We propose a new representation of 3D human shape, equivariant to rotation and invariant to parameterization using the varifolds framework. We propose also a new way to represent a human motion by embedding the 3D shape sequences in infinite dimensional space using a kernel positive definite product from varifolds framework. We compared our method to the combination of dynamic time warping and static human pose descriptors.  Our experiments on 3 datasets showed that our approach gives competitive or better than state-of-the-art results for 3D human motion retrieval, showing better generalization ability than popular deep learning approaches. 
\section*{Acknowledgments}
This work is supported by the ANR project Human4D ANR-19-CE23-0020 and partially by the Investments for the future program  ANR-16-IDEX-0004 ULNE. 




\appendix

\section{Appendix: Comparison with state-of-the-art}

In this section, we explain in more details the state-of-the-art methods. Extensive comparison has been made in~\cite{veinidis2019effective,pierson2021projection} to evaluate the descriptors for human motion retrieval on CVSSP3D dataset. The polygonal curves of those descriptors are filtered with a temporal filtering approach (a mean filter is applied along a temporal window of size $K$). Finally, the dynamic time warping distance is used for comparing the resulting curves. 
We compare our motion retrieval approach to the best features presented in those papers, and to several other learned descriptors:
\begin{enumerate}
    \item The 3D harmonics descriptor~\cite{papadakis20083d}\cite{veinidis2019effective} is a descriptor based on point cloud repartition in space. A 3D shape is first normalized with two variations of PCA. Then, a spherical histogram with different rays is built. The final descriptor is decomposed along spherical harmonics of the obtained with a specific re-weighting for better results. Temporal filtering is proposed in order to deal with the real dataset. We report the results from~\cite{veinidis2019effective}.
    \item Breadths spectrum and shape invariant~\cite{pierson2021projection} are presented as 2 fully invariant descriptors derived from convex shape analysis. The authors propose to use the breadths of the projection of a shape along each axis spanned by a normal $u \in \mathbb{S}^2$ and to keep the rotation invariant spherical spectrum as a descriptor for human pose. They combine the proposed descriptor with weighted areas of the projection on each plan spanned by $u$ to build a shape invariant. Noise robust version of this descriptor, along with specific temporal filtering named Q-breadths and Q-shape invariant are proposed for the real dataset.
    \item Areas, Breadths are the full spherical signals of breadths and weighted areas is proposed to deal with dataset that shows no rotations in~\cite{pierson2021projection}. We apply those descriptors, of size 64, along with their concatenation, Areas \& Breadths, on Dyna dataset.
    \item Aumentado-Armstrong \etal~\cite{gdvae_2019}~propose a variational autoencoder called Geometrically Disentangled VAE (GDVAE). They use PointNet architecture as point cloud encoders and decoders. In the paper, the authors propose to use disentangled intrinsic and extrinsic latent vectors for human shape representation. PointNet encoder is parameterization invariant, but training loss uses the mesh Laplace Beltrami operator which needs a constant parameterization along the training set. Constraints are applied in training to make the network rotation invariant. We report the result of their extrinsic latent vectors (belonging to $\mathbb{R}^{12}$)  from~\cite{pierson2021projection}. The network was pretrained on the SURREAL dataset~\cite{varol17_surreal}. For the CVSSP3D datasets, we report the results from~\cite{pierson2021projection}.
    \item Zhou~\etal~\cite{zhou20unsupervised} propose a mesh autoencoder based on the Neural3DMM~\cite{neural3dmm_2019} mesh neural network architecture. The network is only applied on human shapes, with the objective to disentangle shape and pose in latent space. The network architecture requires that all input meshes have the same parameterization. We can thus apply it only on the artificial dataset. We report the cross validated results from~\cite{pierson2021projection} using the pose latent vectors (belonging to $\mathbb{R}^{112}$) in the human sequence retrieval pipeline. Since  the input of the network are the coordinates of the vertices, the approach is not rotation invariant. For the artificial dataset, we report the results from~\cite{pierson2021projection}.
    \item Cosmo \etal~\cite{cosmo2020limp} propose a similar approach as GDVAE, called Latent Interpolation with Metric Priors (LIMP). They use the same type of autoencoder as GDVAE but change the disentanglement constraints with metric prior constraints: a change in extrinsic latent space should only induce change on extrinsic distances of the meshes, while a change in intrinsic latent space should only induce change on intrinsic distances of the meshes. They use Euclidean and geodesic pairwise matrices in their losses to model this constraint, which needs a constant parameterization in the training set. We use the network pretrained on the FAUST dataset~\cite{Bogo:CVPR:2014}. They do not make any specific training for Euclidean invariance. In order to do motion retrieval, we applied the meshes as input of their available trained network and gathered their extrinsic latent vectors (belonging to $\mathbb{R}^{64}$), and used them in the human sequence retrieval pipeline.
    \item Skinned Multi-Person Linear model (SMPL) pose representation. The SMPL body model~\cite{SMPL:2015} is a parameterized human body model. A template is deformed (non-rigidly) according to a deformation parameterized by a shape vector. A skeleton is associated to this template and a pose vector, composed of relative rotation of each skeletal joint compared to its parent joint. We convert each joint rotation to quaternion representation as in~\cite{zhou20unsupervised,gdvae_2019} and measure the distance between unit quaternions by $d(q, q') = 1-|q.q'|$. The SMPL body pose vector contains the pose information of 20 joints, and the rotation of the central joint accounts for the global rotation of the shape, resulting in a $(\mathbb{R}^4)^{20} = \mathbb{R}^{80}$ representation. Due to the construction of the pose vector, this descriptor is rotation invariant. The SMPL parameters were augmented with dynamic soft tissue deformation relative to each motion (called DMPL) and use to transform the original Dyna dataset to the DFAUST dataset, with better correspondance with the scan. They use for this goal much more information such as texture information from body videos, and the shape vector is retrieved using gender information. We prefer comparing on Dyna dataset rather than DFAUST dataset, allowing us to compare faithfully to the SMPL body pose descriptor. In order to build the pose vectors, a costly fitting method is used along each sequence (accounting in minutes for a single shape). The pose vectors for 129 motions of Dyna where the fitting was successful, we added the SMPL Pose vector retrieved using available code~\url{https://github.com/vchoutas/smplx/} for the remaining 5 motions.
    
\end{enumerate}
\section{Comparison of SPD metrics for Gram-Hankel matrices}
This section is dedicated to the comparison between  Frobenius and Log Euclidean Riemannian Metric (LERM). The Gram-Hankel matrices are positive semidefinite matrices. Several metrics have been propose to compare positive semidefinite matrices. ~\Cref{tab:lerm} shows the results of the comparison between Log Euclidean Riemannian Metric (LERM) and the Frobenius distance.
\begin{equation*}
    d_{LERM}(G_1, G_2) = ||\log(G_1) - \log(G_2)||_F,
\end{equation*}
where $\log (G) = P^T \log(\lambda) P$, where $G = P^T \lambda P$ is the eigen decomposition of the symmetric matrix $G$.
\begin{table*}[]
\centering
\scalebox{0.95}{
\begin{tabular}{l|l|ccc|ccc|ccc}
\multirow{2}{*}{Representation} & \multirow{2}{*}{Gram-Hankel distance} & \multicolumn{3}{l|}{Artificial dataset} & \multicolumn{3}{l|}{Real dataset} & \multicolumn{3}{l}{Dyna dataset} \\ \cline{3-11}
& & NN & FT & ST & NN & FT & ST & NN & FT & ST\\ \hline 
\multicolumn{1}{c|}{\multirow{2}{*}{Current}} & Frobenius & 100 & 100 & 100 & 92.5 & 66.0 & 78.5 & 59.0 & 34.1 & 50.4 \\
& LERM & 100 & 100 & 100 & 78.8 & 55.0 & 76.6 & 55.2 & 35.9 & 51.4 \\ \hline
\multicolumn{1}{c|}{\multirow{2}{*}{Absolute varifolds}} & Frobenius & 100 & 100 & 100 & 95.0 & 66.6 & 80.7 & 60.4 & 40.0 & 55.9  \\
& LERM & 100 & 100 & 100 & 80.0 & 54.6 & 73.4 & 57.5 & 36.0 & 50.8  \\ \hline 
\multicolumn{1}{c|}{\multirow{2}{*}{Oriented varifolds}} & Frobenius & 100 & 100 & 100 & 93.8 & 65.4 & 78.2 & 60.4 & 40.8 & 55.9 \\
& LERM & 100 & 100 & 100 & 86.3 & 50.0 & 66.4  & 57.5 & 37.0 & 51.3 \\

\end{tabular}}
\caption{Motion retrieval results for our approach with Log Euclidean Riemannian Metric (LERM). The results are displayed for CVSSP3D artificial and real datasets, and Dyna datasets}
\label{tab:lerm}
\end{table*}
We observe that the performance is lower than using the Frobenius metric. This results confirms our choice of using Frobenius than LREM metric.
\section{Extended discussion on the parameters $r$ and $\sigma$}
\textbf{Effect of the sigma parameter.}
The performance relative to the $\sigma$ parameter is displayed on the right of Figure 8 in the main paper for oriented varifolds on Dyna dataset. We observe first that the choice of $\sigma$ has a significant impact on performance for NN and in the same time that the optimal $\sigma$ for the NN is not the same as one for FT and ST, for a loss of around $2\%$ in those metrics, which is less significant than the NN gain. \\
\textbf{Effect of the choice of r.}
The performance relative to the $r$ parameter is displayed on the left of Figure 8 for oriented varifolds on Dyna dataset. We observe first that the choice of $r$ has a significant impact on performance and in the same time that the optimal $r$ for the NN is not the same as one for FT and ST, for a loss of around $5\%$ in those metrics.\\
\textbf{Effect of normalizations.}
We present in Table 2 of the main paper 
the performances of oriented varifolds with the 2 normalization techniques presented here. The centroid normalization is essential to the good performance of our approach. In the mean time, the inner product normalization always implies significant boost for NN metric, but can induce a (non-significant) loss in ST and FT metrics. 

\section{Qualitative results: Queries on Dyna}
\begin{figure}
    \includegraphics{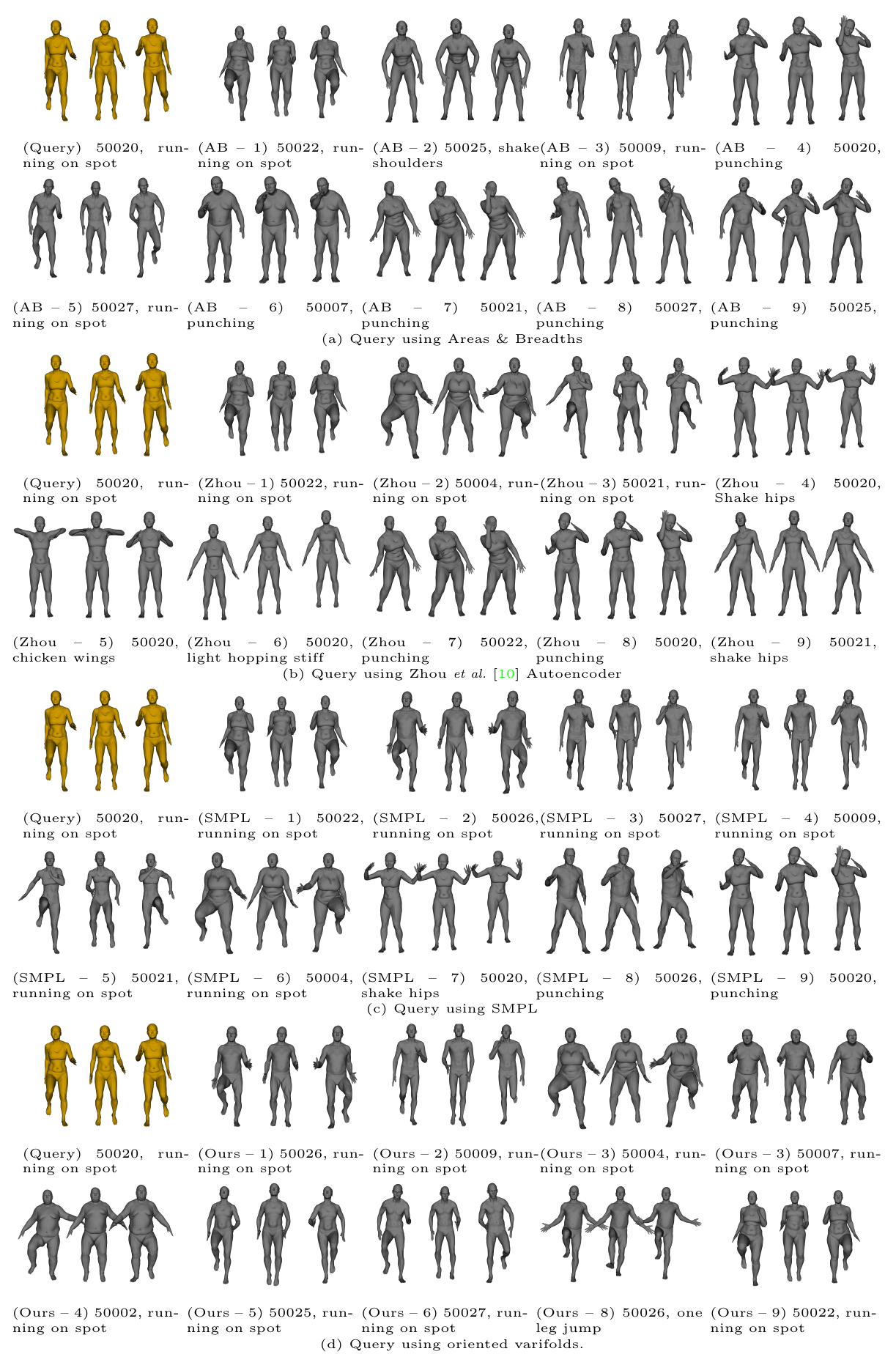}
    \caption{First tier of the query of the paper for Areas \& Breadths~\cite{pierson2021projection}, Zhou~\etal~\cite{zhou20unsupervised}, SMPL~\cite{Bogo:CVPR:2014} and oriented varifolds on the Dyna dataset. The query is in yellow and the results are sorted by closeness to the query using a given approach.}
    \label{fig:all_queries}
\end{figure}

 \Cref{fig:all_queries} shows the results for SMPL, Zhou et al and Areas \& Breadths. Although it is a the first tier is better for our approach in two manners: First we observe that there is no confusion between a motion and the motion of the same individual in our approach. Secondly, some drawbacks of the other methods appear: Areas \& Breadths are symmetric descriptors and does not make the difference between a punching arm (from down to up) and the two arms that goes up and down when running, and we see a lot of punching motions retrieved (4 out of 6 wrong retrievals).  Second, the autoencoder of Zhou~\etal is not fully disentangled from the identity of the body and a lot of motions from the same identity are retrieved (4 out of 6 wrong retrievals). SMPL gives the best result, as expected from Table 1 of the paper. However, we observe also some sensitivity to the identity of the performer (2 out of 3 wrong retrievals).

\section{Qualitative results on CVSSP3D real dataset.}
\begin{figure}
        \centering
        \includegraphics[width=0.50\linewidth]{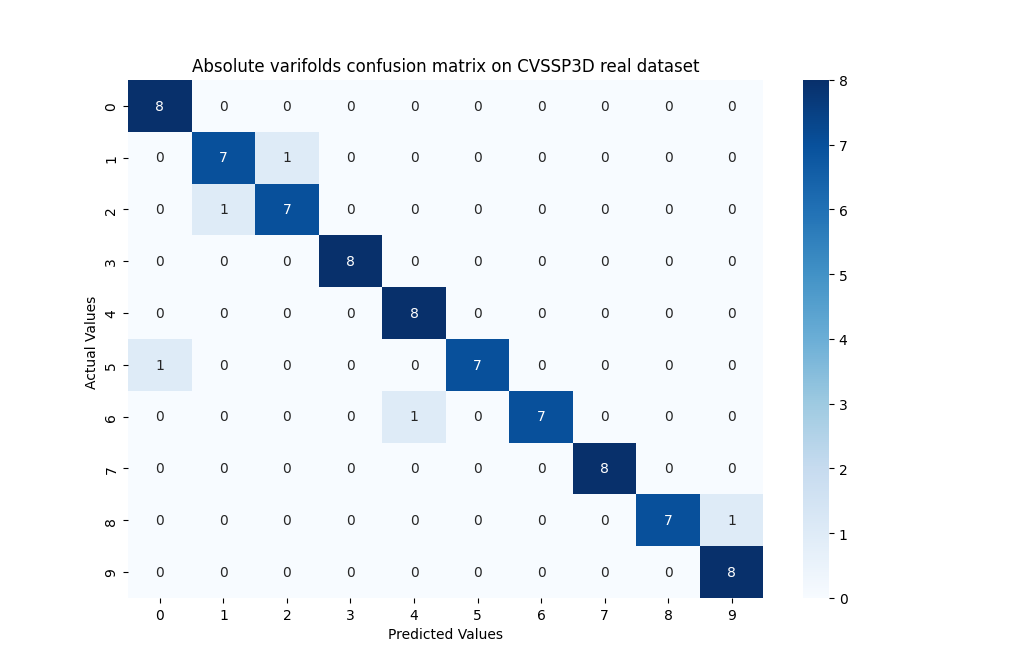}
        \caption{NN Confusion matrix of absolute varifolds on CVSSP3D real dataset.}
        \label{fig:conf_real}
        \vspace{-5mm}
    \end{figure}

In the CVSSP3D real dataset, clothes worn by the subjects during the acquisition process induce topological and mesh noises (see Figure 1 and Figure 5(b) of the paper). The results on this dataset shows our method robustness to the noise and clothes present in clothed human dataset.
The quantitative results in Table 1 (paper) show that our approach is robust to the noise and outperforms state-of-art methods on CVSSP3D real dataset in terms of NN. The confusion matrix of our approach (absolute varifolds) on CVSSP3D real dataset~, in~\Cref{fig:conf_real} shows that our approach performs well on all human motions of the dataset. We display also a query with absolute varifolds, in~\Cref{fig:query} (same query as the one displayed in~\cite{pierson2021projection}). Our approach is able to provide 6 out of the 7 walk motion, showing a slighlty better results compared to~\cite{pierson2021projection} (5 out of 7).

\begin{figure}
    \includegraphics{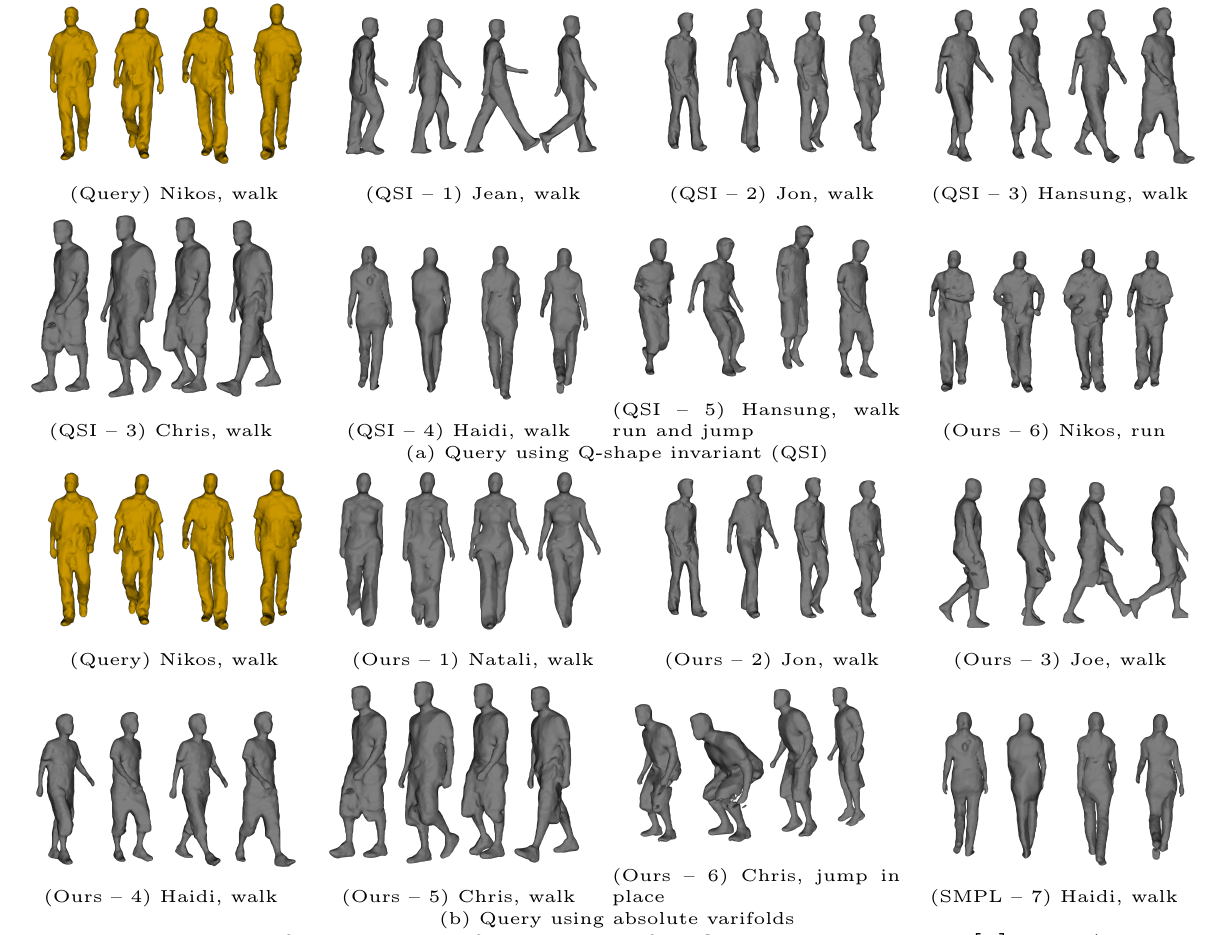}
    \caption{First tier of the query of the paper for Q-shape invariant~\cite{pierson2021projection} and Absolute Varifolds. The query is in yellow and the results are sorted by closeness to the query using a given approach. The first query is directly taken from~\cite{pierson2021projection}. The query is in yellow and the results are sorted by closeness to the query using a given approach.}
    \label{fig:query}
\end{figure}

\end{document}